\documentclass[final]{cvpr}

\PassOptionsToPackage{numbers}{natbib}

\usepackage{times}
\usepackage{epsfig}
\usepackage{graphicx}
\usepackage{amsmath}
\usepackage{amssymb}

\usepackage{bm}
\usepackage{algorithm}
\usepackage{algpseudocode}
\usepackage{mathtools} %
\usepackage{dsfont} %
\usepackage{bbm}
\usepackage{booktabs} %
\usepackage{diagbox} 
\usepackage{nccmath} %
\usepackage{arydshln} %
\usepackage{tablefootnote}
\usepackage{bbding} %
\usepackage{amssymb}
\usepackage{enumitem} %
\usepackage{subfigure}
\usepackage{amsmath}

\usepackage{xcolor}
\usepackage{caption}
\usepackage{wrapfig}
\usepackage{multirow}

\usepackage{soul}

\usepackage{stackengine}

\newcommand{\TE}{T\!E}

\newcommand{\NDE}{N\!D\!E}
\newcommand{\NIE}{N\!I\!E}
\newcommand{\TDE}{T\!D\!E}
\newcommand{\TIE}{T\!I\!E}
\newcommand{\softmax}{\text{softmax}}

\newcommand{\bfit}[1]{\textbf{\textit{{#1}}}}

\renewcommand{\caption}[1]{\vspace{-3mm}\caption{#1}\vspace{-3mm}}

\usepackage[pagebackref=true,breaklinks=true,colorlinks,bookmarks=true]{hyperref}

\def\cvprPaperID{****} %
\def\confYear{CVPR 2021}

\begin{document}

\title{Counterfactual VQA: A Cause-Effect Look at Language Bias}

\author{
Yulei Niu$^{1}$~~~
Kaihua Tang$^{1}$~~~
Hanwang Zhang$^{1}$~~~
Zhiwu Lu$^{2,3}$~~~
Xian-Sheng Hua$^{4}$~~~
Ji-Rong Wen$^{2,3}$\\
\small $^{1}$Nanyang Technological University, Singapore $^{2}$Gaoling School of Artificial Intelligence, Renmin University of China, Beijing, China\\
\small $^{3}$Beijing Key Laboratory of Big Data Management and Analysis Methods $^{4}$Damo Academy, Alibaba Group\\
{\tt\small yn.yuleiniu@gmail.com, \{kaihua001@e., hanwangzhang@\}ntu.edu.sg, \{luzhiwu, jrwen\}@ruc.edu.cn}}

\maketitle

\begin{abstract}
VQA models may tend to rely on language bias as a shortcut and thus fail to sufficiently learn the multi-modal knowledge from both vision and language. Recent debiasing methods proposed to exclude the language prior during inference. However, they fail to disentangle the ``good'' language context and ``bad'' language bias from the whole. In this paper, we investigate how to mitigate language bias in VQA. Motivated by causal effects, we proposed a novel counterfactual inference framework, which enables us to capture the language bias as the direct causal effect of questions on answers and reduce the language bias by subtracting the direct language effect from the total causal effect. Experiments demonstrate that our proposed counterfactual inference framework 1) is general to various VQA backbones and fusion strategies, 2) achieves competitive performance on the language-bias sensitive VQA-CP dataset while performs robustly on the balanced VQA v2 dataset without any augmented data. The code is available at \url{https://github.com/yuleiniu/cfvqa}. 
\end{abstract}
\section{Introduction}\label{sec:intro}

Visual Question Answering (VQA)~\cite{antol2015vqa,agrawal2017vqa} has become the fundamental building block that underpins many frontier interactive AI systems, such as visual dialog~\cite{das2017visual}, vision-and-language navigation~\cite{anderson2018vision}, and visual commonsense reasoning~\cite{zellers2019recognition}. 
VQA systems are required to perform visual analysis, language understanding, and multi-modal reasoning.
Recent studies~\cite{goyal2017making,agrawal2018don,antol2015vqa,goyal2017making,kafle2017analysis} found that VQA models may rely on spurious linguistic correlations rather than multi-modal reasoning.
For instance, simply answering ``tennis'' to the sport-related questions and ``yes'' to the questions ``Do you see a ...'' can achieve approximately 40\% and 90\% accuracy on the VQA v1.0 dataset. 
As a result, VQA models will fail to generalize well if they simply memorize the strong language priors in the training data~\cite{agrawal2016analyzing,goyal2017making}, especially on the recently proposed VQA-CP ~\cite{agrawal2018don} dataset where 
the priors are quite different in the training and test sets.

One straightforward solution to mitigate language bias is to enhance the training data by using extra annotations or data augmentation. In particular, visual~\cite{das2017human} and textual~\cite{huk2018multimodal} explanations are used to improve the visual grounding ability~\cite{selvaraju2019taking,wu2019self}. Besides, counterfactual training samples generation~\cite{chen2020counterfactual,abbasnejad2020counterfactual,zhu2020overcoming,gokhale2020mutant,liang2020learning} helps to balance the training data, and outperform other debiasing methods by large margins on VQA-CP.
These methods demonstrate the effect of \textit{debiased training} to improve the generalizability of VQA models. However, it is worth noting that VQA-CP was proposed to validate whether VQA models can disentangle the learned visual knowledge and memorized language priors~\cite{agrawal2018don}. Therefore, how to make unbiased inference under \textit{biased training} still remains a major challenge in VQA. Another popular solution~\cite{cadene2019rubi,clark2019don} is using a separate question-only branch to learn the language prior in the training set. During the test stage, the prior is mitigated by excluding the extra branch. However, we argue that the language prior consists of both ``bad'' language bias (\eg, binding the color of bananas with the major color ``yellow'')
and ``good'' language context (\eg, narrowing the answer space based on the question type ``what color''). 
Simply excluding the extra branch cannot make use of the good context. Indeed, it is still challenging for recent debiasing VQA methods to disentangle the good and bad from the whole.

\begin{figure*}
    \centering
    \includegraphics[width=0.95\linewidth]{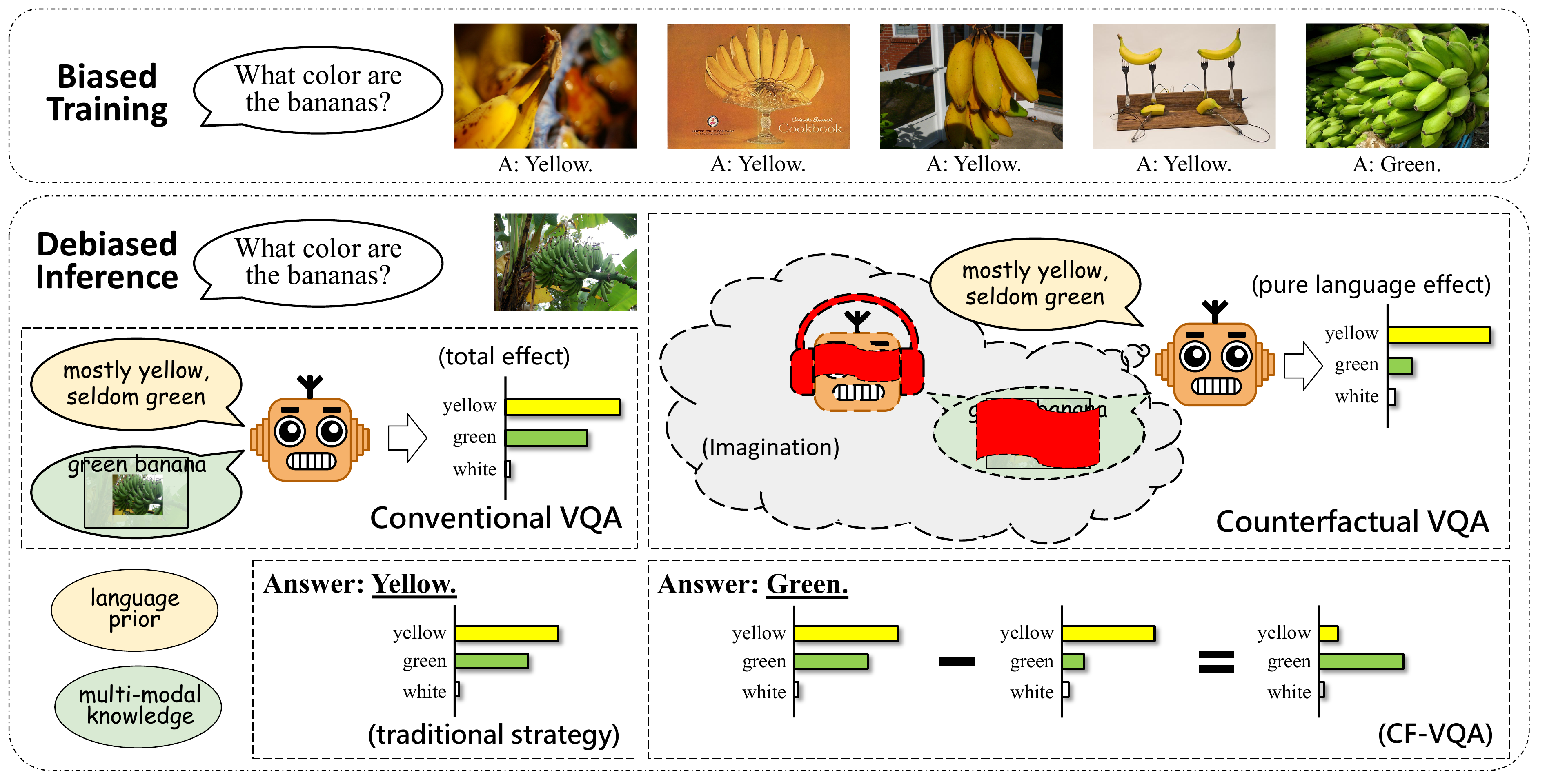}
    \vspace{-2mm}
    \caption{Our cause-effect look at language bias in VQA. Conventional VQA depicts the fact where machine hears the question and extracts the multi-modal knowledge. Counterfactual VQA depicts the scenario where machine hears the question but the knowledge is blocked. We subtract the pure language effect from the total effect for debiased inference.} 
    \vspace{-3mm}
    \label{fig:teaser}
\end{figure*}

Motivated by counterfactual reasoning and causal effects~\cite{pearl2000causality,pearl2001direct,pearl2018book}, we propose a novel counterfactual inference framework called CF-VQA to reduce language bias in VQA. Overall, we formulate language bias as the direct causal effect of questions on answers, and mitigate the bias by subtracting the direct language effect from the total causal effect. As illustrated in Figure~\ref{fig:teaser}, we introduce two scenarios, conventional VQA and counterfactual VQA, to estimate the total causal effect and direct language effect, respectively. These two scenarios are defined as follows:

\noindent \textbf{Conventional VQA}: \textit{What will answer $A$ be, if machine hears question $Q$, sees image $V$, and extracts the multi-modal knowledge $K$?}

\noindent \textbf{Counterfactual VQA}:  \textit{What would $A$ be, if machine hears $Q$, but had not extracted $K$ or seen $V$?}

\noindent Intuitively, conventional VQA depicts the scenario where 
both $Q$ and $V$ are available. In this case, we can estimate the total causal effect of $V$ and $Q$ on $A$. However, conventional VQA cannot disentangle the single-modal linguistic correlation and multi-modal reasoning, \ie, direct and indirect effects. Therefore, we consider the following counterfactual question: \textit{``What would have happened if the machine had not performed multi-modal reasoning?''} The answer to this question can be obtained by imagining a scenario where the machine hears $Q$, but the multi-modal knowledge $K$ is intervened under the no-treatment condition, \ie, $V$ and $Q$ had not been accessible. 
Since the response of $K$ to $Q$ is blocked, VQA models can only rely on the single-modal impact. Therefore, language bias can be identified by estimating the direct causal effect of $Q$ on $A$, \ie, pure language effect.
The training stages follows language-prior based methods~\cite{cadene2019rubi,clark2019don} that train an ensemble model with a prevailing VQA model and single-modal branches. During the test stage, CF-VQA uses the debiased causal effect for inference, which is obtained by subtracting the pure language effect from the total effect. 
Perhaps surprisingly, recent language-prior based methods~\cite{cadene2019rubi,clark2019don} can be further unified into our proposed counterfactual inference framework as special cases. In particular, CV-VQA can easily improve RUBi~\cite{cadene2019rubi} by 7.5\% with only one more learnable parameter.
Experimental results show that CF-VQA outperforms the methods without data argumentation by large margins on the VQA-CP dataset~\cite{agrawal2018don} while remaining stable on the balanced VQA v2 dataset~\cite{goyal2017making}.

The main contribution of this paper is threefold. First, our counterfactual inference framework is the first to formulate language bias in VQA as causal effects. Second, we provide a novel causality-based interpretation for recent debiasing VQA works~\cite{cadene2019rubi,clark2019don}. Third, our cause-effect look is general and suitable for different baseline VQA architectures and fusion strategies.

\section{Related Work}
\noindent \textbf{Language Bias in VQA} can be interpreted in two ways. First, there exists strong correlations between questions and answers, which reflects the \textit{``language prior''}~\cite{goyal2017making,agrawal2018don}. Simply answering ``tennis'' to the sport-related questions can achieve approximately 40\% accuracy on the VQA v1.0 dataset. Second, 
the questioner tends to ask about the objects seen in the image, which leads to the \textit{``visual priming bias''}~\cite{antol2015vqa,goyal2017making,kafle2017analysis}. Simply answering ``yes'' to all the questions ``Do you see a ...'' achieves nearly 90\% accuracy on the VQA v1.0 dataset. In both ways, machines may merely focus on the question rather than the visual content. 
This serious shortcut limits the generalization of VQA models
~\cite{agrawal2016analyzing,zhang2016yin,goyal2017making}, especially when the test scenario is quite different from the training scenario. 

\noindent \textbf{Debiasing Strategies in VQA.} Recently, a new VQA split, Visual Question Answering under Changing Priors (VQA-CP)~\cite{agrawal2018don}, was proposed to evaluate the generalizability of VQA models. In VQA-CP, the distributions of answers for every question type are different during training and test stages. Most of recent solutions to reduce the language bias in VQA can be grouped into three categories, strengthening visual grounding~\cite{selvaraju2019taking,wu2019self}, weakening language prior~\cite{ramakrishnan2018overcoming,cadene2019rubi,clark2019don}, and implicit/explicit data argumentation~\cite{chen2020counterfactual,abbasnejad2020counterfactual,teney2020unshuffling,zhu2020overcoming}. First, human visual~\cite{das2017human} and textual~\cite{huk2018multimodal} explanations are exploited to strengthen the visual grounding in VQA~\cite{selvaraju2019taking,wu2019self}. 
Second, ensemble-based methods proposed to use a separated QA branch to capture the language prior under adversarial learning~\cite{ramakrishnan2018overcoming} or multi-task learning~\cite{cadene2019rubi,clark2019don}. 
Third, recent works~\cite{chen2020counterfactual,abbasnejad2020counterfactual} automatically generate additional question-image pairs to balance the distribution of training data. 
In this paper, the language-prior based methods~\cite{cadene2019rubi,clark2019don} can be unified into our proposed counterfactual inference framework as special cases.

\noindent \textbf{Causality-inspired Computer Vision.}
Counterfactual thinking and causal inference have inspired several studies in computer vision, including visual explanations~\cite{goyal2019counterfactual,anne2018grounding,wang2020scout,yi2019clevrer}, scene graph generation~\cite{chen2018scene,tang2020unbiased}, image recognition~\cite{tang2020unbiased}, video analysis~\cite{fang2019modularized,kanehira2019multimodal}, zero-shot and few-shot learning~\cite{yue2020interventional,yue2021counterfactual} incremental learning~\cite{hu2021distilling}, representation learning~\cite{wang2020visual,zhang2020devlbert}, semantic segmentation~\cite{zhang2020causal}, and vision-language tasks~\cite{chen2020counterfactual,teney2020learning,qi2019two,yang2020deconfounded,fu2020sscr,yang2021causal}. 
Especailly, counterfactual learning has been exploited in recent VQA studies~\cite{chen2020counterfactual,teney2020learning,abbasnejad2020counterfactual}. Different from these works that generate counterfactual  samples for \textit{debiased} training, our cause-effect look focuses on counterfactual inference with even \textit{biased} training data. 

\section{Preliminaries}\label{sec:pre}
In this section, we introduce the used concepts of causal inference~\cite{pearl2000causality,pearl2001direct,robins2003semantics,pearl2018book}. In the following, we represent a random variable as a capital letter (\eg, $X$), and denote its observed value as a lowercase letter (\eg, $x$).  

\begin{figure}
\centering
{
\subfigure[Causal Graph]{
\label{fig:2-causal-graph-a}
\includegraphics[width=0.31\linewidth]{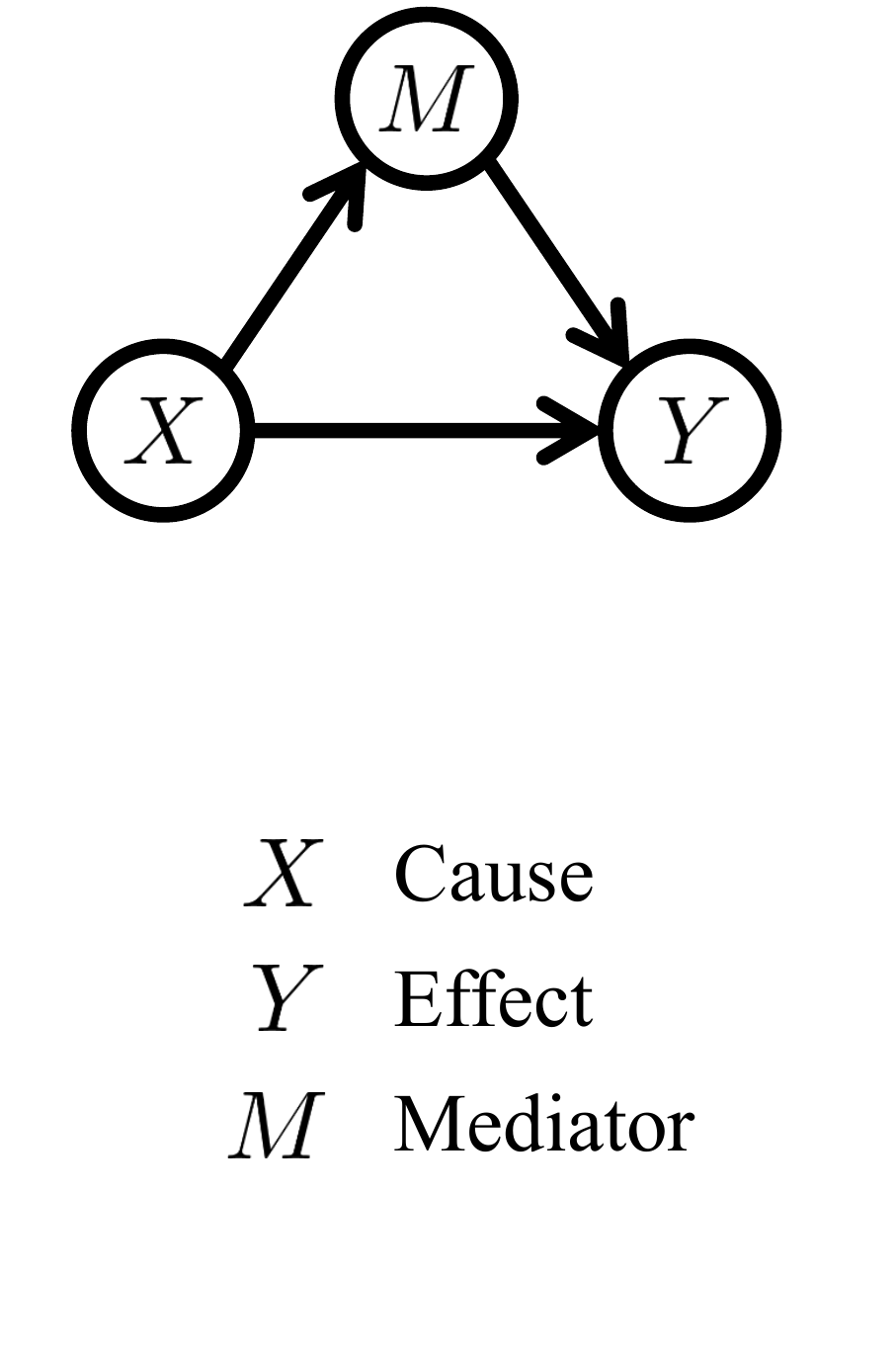}
}
\subfigure[Counterfactual Notations]{
\label{fig:2-causal-graph-b}
\includegraphics[width=0.6\linewidth]{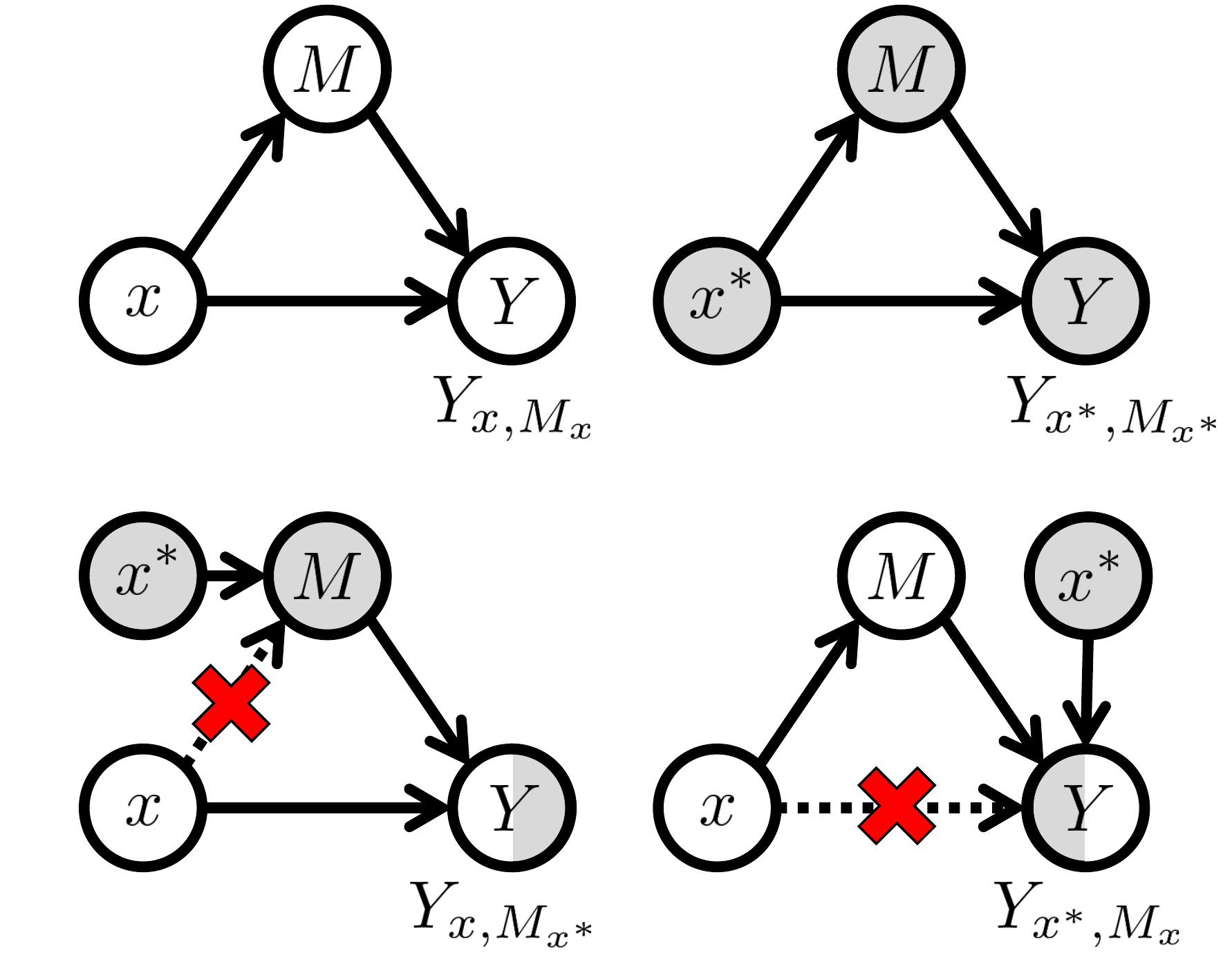}
}
}
\vspace{-2mm}
\caption{(a) Example of causal graph. (b) Examples of counterfactual notations. White nodes are at the value $X\!=\!x$ while gray nodes are at the value $X\!=\!x^*$.}
\label{fig:2-causal-graph}
\end{figure}

\noindent \textbf{Causal graph} reflects the causal relationships between variables, which is represented as a directed acyclic graph $\mathcal{G}=\{\mathcal{V},\mathcal{E}\}$, where $\mathcal{V}$ denotes the set of variables and $\mathcal{E}$ represents the cause-and-effect relationships. Figure \ref{fig:2-causal-graph-a} shows an example of causal graph that consists of three variables.
If the variable $X$ has a \textit{direct} effect on the variable $Y$, we say that 
$Y$ is the child of $X$,
\ie, $X\!\rightarrow\!Y$. 
If $X$ has an \textit{indirect} effect on $Y$ via the variable $M$, we say that $M$ acts as a \textit{mediator} between $X$ and $Y$, \ie, $X\!\rightarrow\!M\!\rightarrow\!Y$. 

\noindent \textbf{Counterfactual notations} are used to translate causal assumptions from graphs to formulas. 
The value that $Y$ would obtain if $X$ is set to $x$ and $M$ is set to $m$ is denoted as:

\begin{equation}\label{eq:cfnot}
Y_{x,m} = Y(X=x,M=m).\footnote{If there is no confounder of $X$, then we have that $do(X\!=\!x)$ is equivalent to $X\!=\!x$ and can omit the $do$ operator.}
\end{equation}
In the factual scenario, we have $m\!=\!M_x\!=\!M(X\!=\!x)$.
In the counterfactual scenario, $X$ is set as different values for $M$ and $Y$. For example, $Y_{x,M_{x^*}}$ describes the situation where $X$ is set to $x$ and $M$ is set to the value when $X$ had been $x^*$, \ie, $Y_{x,M_{x^*}} = Y(X\!=\!x,M\!=\!M(X\!=\!x^*))$.
Note that $X$ can be simultaneously set to different values $x$ and $x^*$ only in the counterfactual world. Figure~\ref{fig:2-causal-graph-b} illustrates examples of counterfactual notations.

\noindent \textbf{Causal effects} reflect the comparisons between two potential outcomes of the same individual given two different treatments~\cite{rubin1978bayesian,robins1986new}.
Supposed that $X=x$ represents ``under treatment condition'' and $X=x^*$ represents ``under no-treatment condition''\footnote{For example, to estimate the effect of a drug on a disease, $X=x$ represents taking the drug, while $X=x^*$ represents not taking the drug.}. The total effect (TE) of treatment $X\!=\!x$ on $Y$ compares two hypothetical situations $X\!=\!x$ and $X\!=\!x^*$, which is denoted as:
\begin{equation}\label{eq:TE}
    \TE = Y_{x,M_x}-Y_{x^*,M_{x^*}}.
\end{equation}
Total effect can be decomposed into natural direct effect (NDE) and total indirect effect (TIE). NDE denotes the effect of $X$ on $Y$ with the mediator $M$ blocked, and expresses the increase in $Y$ with $X$ changing from $x^*$ to $x$,
while $M$ is set to the value it \textit{would have obtained} at $X=x^*$, \ie, the response of $M$ to the treatment $X\!=\!x$ is disabled:
\begin{equation}\label{eq:NDE}
\begin{split}
    \NDE &= Y_{x,M_{x^*}}-Y_{x^*,M_{x^*}}. \\
\end{split}
\end{equation}
TIE is the difference between TE and NDE, denoted as: 
\begin{equation}\label{eq:TIE}
\begin{split}
    \TIE &= \TE-\NDE = Y_{x,M_{x}}-Y_{x,M_{x^*}}.
\end{split}
\end{equation}
TE can be also decomposed into natural indirect effect (NIE) and total direct effect (TDE). Similarly, NIE reflects the effect of $X$ on $Y$ through the mediator $M$, \ie, $X\!\rightarrow\!M\!\rightarrow\!Y$,
while the direct effect on $X\!\rightarrow\!Y$ is blocked by setting $X$ as $x^*$.
NIE is denoted as:
\begin{equation}\label{eq:NIE}
\begin{split}
    \NIE &= Y_{x^*,M_{x}}-Y_{x^*,M_{x^*}}. \\
\end{split}
\end{equation}
In Section~\ref{sec:cfvqa}, we will further discuss the meanings and differences of these effects in VQA.

\section{Cause-Effect Look at VQA}\label{sec:cfvqa}

Following the common formulation, we define the VQA task as a multi-class classification problem. VQA models are required to select an answer from the candidate set $\mathcal{A}=\{a\}$ given an image $V=v$ and a question $Q=q$. 

\subsection{Cause-Effect Look}\label{sec:celook}

\begin{figure}
\centering
{
\subfigure[]{
\label{fig:vqacg-full-1}
\includegraphics[width=0.3\linewidth]{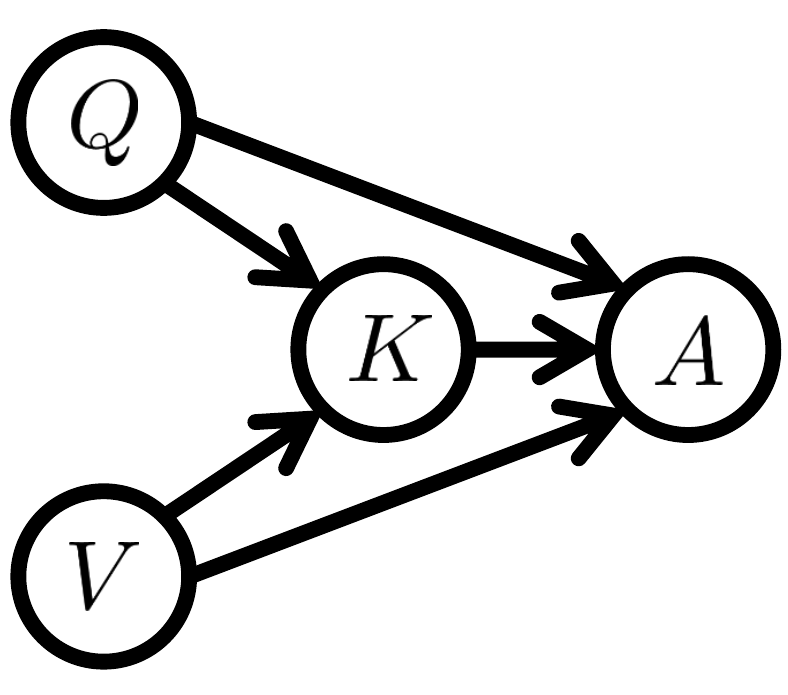}
}
\subfigure[]{
\label{fig:vqacg-full-2}
\includegraphics[width=0.64\linewidth]{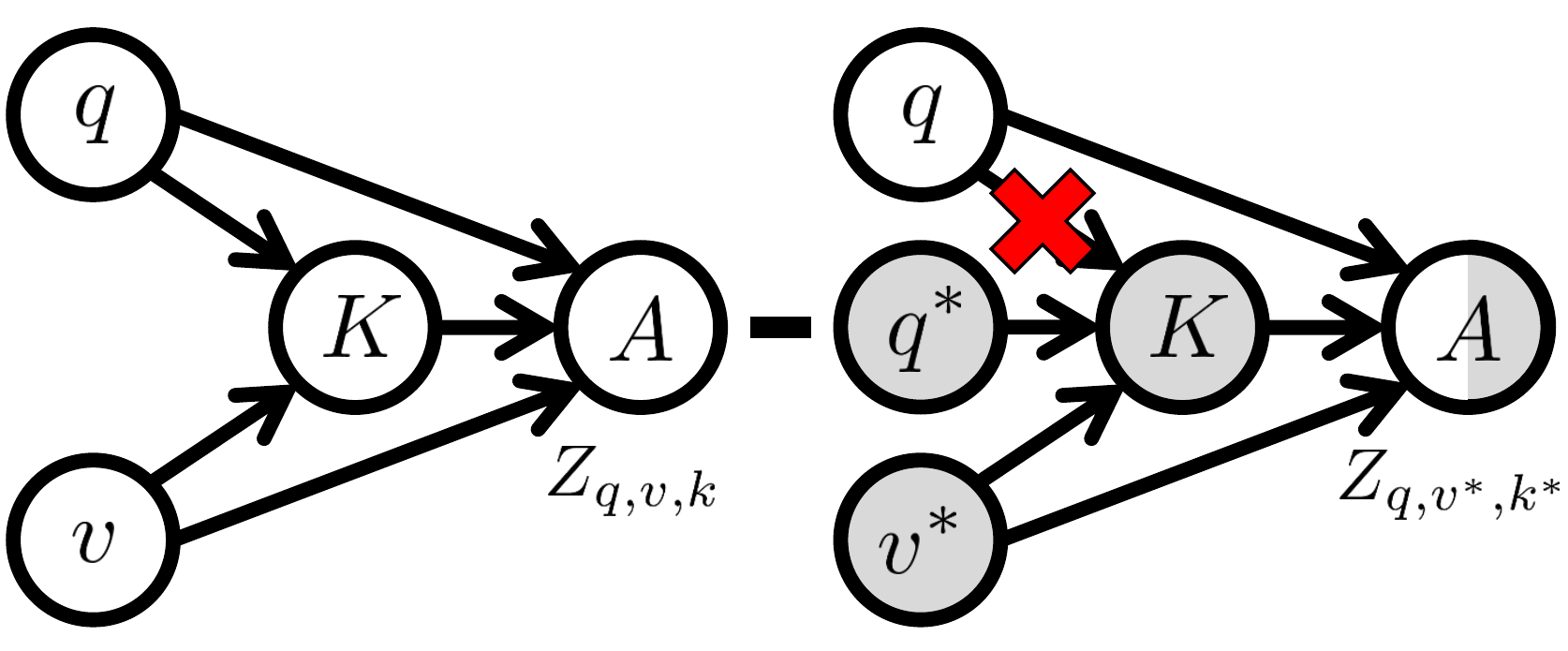}
}
}
\caption{(a) Causal graph for VQA. $Q$: question. $V$: image. $K$: multi-modal knowledge. $A$: answer. (b) Comparison between conventional VQA (left) and counterfactual VQA (right). White nodes are at the value $V\!=\!v$ and $Q\!=\!q$ while gray nodes are at the value $V\!=\!v^*$ and $Q\!=\!q^*$.}
\label{fig:vqacg-full}
\vspace{-3mm}
\end{figure}

The causal graph of VQA is illustrated in Figure~\ref{fig:vqacg-full-1}. The effect of $V$ and $Q$ on $A$ can be divided into the single-modal impact and multi-modal impact. The single-modal impact captures the \textit{direct} effect of $V$ or $Q$ on $A$ via $V\rightarrow A$ or $Q\rightarrow A$. The multi-modal impact captures the \textit{indirect} effect of $V$ and $Q$ on $A$ via the multi-modal knowledge $K$, \ie, $V,Q\rightarrow K \rightarrow A$. 
We propose to exclude pure language effect on $Q\!\rightarrow\!A$ to reduce language bias in VQA.

Following the counterfactual notations in Eq.~\eqref{eq:cfnot}, we denote the score that the answer $a$ (\eg, ``green'') would obtain if $V$ is set to $v$ (\eg, an image which includes green bananas) and $Q$ is set to $q$ (\eg, a question ``What color are the bananas?'') as
\begin{equation*}
    Y_{v,q}(a)=Y(a;V\!=\!v,Q\!=\!q).
\end{equation*}
Without loss of generality, we omit $a$ for simplicity, \ie, $Y_{v,q}=Y(V\!=\!v,Q\!=\!q)$. Similarly, the counterfactual notation of $K$ is denoted as $K_{v,q}=K(V\!=\!v,Q\!=\!q)$.

As shown in Figure~\ref{fig:vqacg-full-1}, there exist three paths directly connected to $A$, \ie, $Q\!\rightarrow\!A$, 
$V\!\rightarrow\!A$, 
and $K\!\rightarrow\!A$. 
Therefore, we rewrite $Y_{v,q}$ as the function of $Q$, $V$ and $K$:
\begin{equation}\label{eq:z}
    Y_{v,q}=Z_{q,v,k}=Z(Q=q,V=v,K=k),
\end{equation}
where $k\!=\!K_{v,q}$. Following the definition of causal effects in Section~\ref{sec:pre}, the \textit{total effect} (TE) of $V\!=\!v$ and $Q\!=\!q$ on $A\!=\!a$ can be written as:
\begin{equation}\label{eq:vqate}
    \TE=Y_{v,q}-Y_{v^*,q^*}=Z_{q,v,k}-Z_{q^*,v^*,k^*},
\end{equation}
where $k^*\!=\!K_{v^*,q^*}$. Here $v^*$ and $q^*$ represent the no-treatment condition where $v$ and $q$ are not given. 

As we have discussed in Section~\ref{sec:intro}, VQA models may suffer from the spurious correlation between questions and answers, and thus fail to conduct effective multi-modal reasoning. Therefore, we expect VQA models to exclude the direct impact of questions. To achieve this goal, we proposed \textit{counterfactual VQA} to estimate the causal effect of $Q\!=\!q$ on $A\!=\!a$ by blocking the effect of $K$ and $V$.
Counterfactual VQA describes the scenario where $Q$ is set to $q$ and $K$ would attain the value $k^*$ when $Q$ had been $q^*$ and $V$ had been $v^*$. Since the response of mediator $K$ to inputs is blocked, the model can only rely on the given question for decision making. Figure~\ref{fig:vqacg-full-2} shows the comparison between conventional VQA and counterfactual VQA. We obtain the \textit{natural direct effect} (NDE) of $Q$ on $A$ by comparing counterfactual VQA to the no-treatment conditions:
\begin{equation}\label{eq:vqande}
    \NDE=Z_{q,v^*,k^*}-Z_{q^*,v^*,k^*}.
\end{equation}
Since the effect of $Q$ on the intermediate $K$ is blocked (\ie, $K\!=\!k^*$), 
NDE explicitly captures the language bias. Furthermore, the reduction of language bias can be realized by subtracting NDE from TE, which is represented as:
\begin{equation}\label{eq:vqatie}
    \TIE=\TE-\NDE=Z_{q,v,k}-Z_{q,v^*,k^*}.
\end{equation}
We select the answer with the maximum TIE for inference, which is totally different from traditional strategies that is based on the posterior probability 
 \ie, $P(a|v,q)$. 

\subsection{Implementation}\label{sec:imp}
\noindent \textbf{Parameterization}. The calculation of the score $Z_{q,v,k}$ in Eq.~\eqref{eq:z} is parameterized by three neural models $\mathcal{F}_Q$, $\mathcal{F}_V$, $\mathcal{F}_{VQ}$ and one fusion function $h$ as:
\begin{equation}\label{eq:archi}
\begin{gathered}
Z_q=\mathcal{F}_Q(q),\quad Z_v=\mathcal{F}_V(v),\quad Z_k=\mathcal{F}_{VQ}(v,q),\\
Z_{q,v,k}=h(Z_q,Z_v,Z_k),
\end{gathered}
\end{equation}
where $\mathcal{F}_Q$ is the language-only branch (\ie, $Q\!\rightarrow\!A$), 
$\mathcal{F}_V$ is the vision-only branch (\ie, $V\!\rightarrow\!A$), 
and $\mathcal{F}_{VQ}$ is the vision-language branch (\ie, $V,Q\!\rightarrow\!K\!\rightarrow\!A$).
The output scores are fused by the function $h$
to obtain the final score $Z_{q,v,k}$. 

\begin{figure*}
    \centering
    \includegraphics[width=1\linewidth]{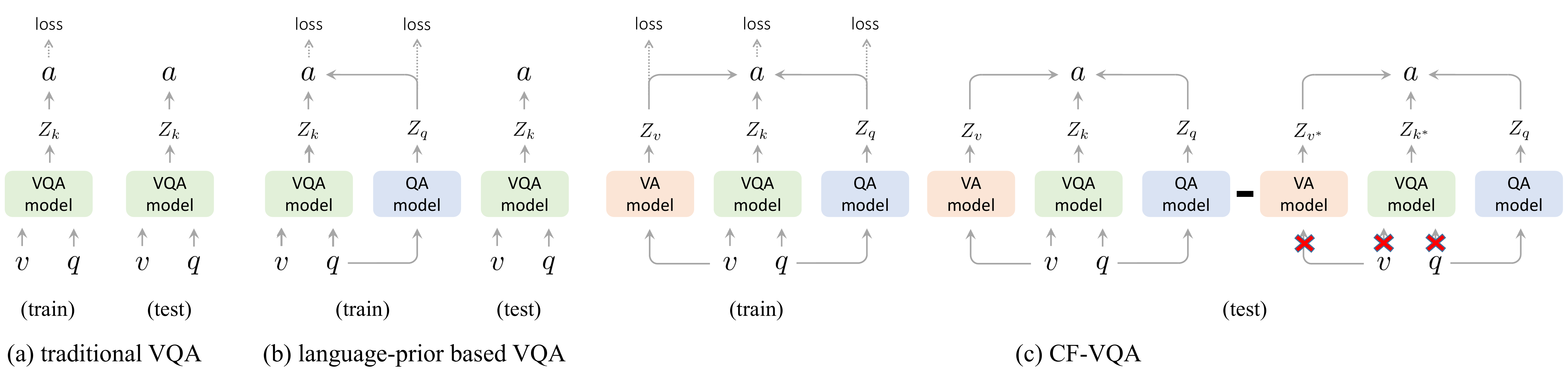}
    \caption{Comparison between our CF-VQA and other VQA methods. (a) Traditional VQA methods use a single VQA model. (b) Language-prior based methods~\cite{cadene2019rubi,clark2019don} use an additional QA model to capture the language prior during the training stage. The QA model is not used during testing. (c) CF-VQA maintains the QA model during the test stage, and makes inference based on the debiased causal effect. The VA model is optional depending on the causal graph assumption.}
    \label{fig:comp}
\end{figure*}

As introduced in Section~\ref{sec:celook}, the no-treatment condition is defined as blocking the signal from vision or language, \ie, $v$ or $q$ is not given. We represent the no-treatment conditions as $V\!=\!v^*\!=\!\varnothing$ and $Q\!=\!q^*\!=\!\varnothing$. 
Note that neural models cannot deal with no-treatment conditions where the inputs are void.
Therefore, we assume that the model will randomly guess with equal probability under the no-treatment conditions. In this case, $Z_q$, $Z_v$ and $Z_k$ in Eq.~\eqref{eq:archi} can be represented as:
\begin{equation}
\begin{split}
    Z_q=
    \begin{cases}
    z_{q}=\mathcal{F}_{Q}(q) & \text{ if $Q=q$}\\
    z^*_{q}=c & \text{ if $Q=\varnothing$}\\
    \end{cases},
\end{split}\label{eq:zq}
\end{equation}
\begin{equation}
\begin{split}
    Z_v=
    \begin{cases}
    z_{v}=\mathcal{F}_{V}(v) & \text{ if $V=v$}\\
    z^*_{v}=c & \text{ if $V=\varnothing$}\\
    \end{cases},
\end{split}\label{eq:zv}
\end{equation}
\begin{equation}
\begin{split}
    Z_{k}=
    \begin{cases}
    z_k=\mathcal{F}_{VQ}(v,q) & \text{ if $V=v$ and $Q=q$}\\
    z^*_k=c & \text{ if $V=\varnothing$ or $Q=\varnothing$}\\
    \end{cases},
\end{split}\label{eq:zk}
\end{equation}
where $c$ denotes a learnable parameter. We use the uniform distribution assumption for two reasons. Firstly, as for us human, we would like to make a wild guess if we have absolutely no idea about the specific treatments, including question types or topics. 
Secondly, as $z^*_{v}$ and $z^*_{k}$ are used to estimated NDE of $Q$, the uniform distribution can guarantee a safe estimation. We further empirically validate the assumption in the ablation study.

\noindent \textbf{Fusion Strategies}. 
We expect that the fused score $Z_{q,v,k}$ is a combination of $Z_q$, $Z_v$ and $Z_k$.
We proposed two non-linear fusion variants, Harmonic (HM) and SUM:
\begin{flalign}\label{eq:fusion-har}
\text{(HM)}&&\mkern4mu h(Z_q,Z_v,Z_k)=\log\frac{Z_{\text{HM}}}{1+Z_{\text{HM}}},&&
\end{flalign}
where $Z_{\text{HM}}\!=\!\sigma(Z_q)\cdot\sigma(Z_v)\cdot\sigma(Z_k)$.
\begin{flalign}\label{eq:fusion-sum}
\text{(SUM)}&&h(Z_q,Z_v,Z_k)=\log\sigma(Z_{\text{SUM}}),&&
\end{flalign}
where $Z_{\text{SUM}}=Z_q+Z_v+Z_k$.

\noindent \textbf{Training}. The training strategy follows~\cite{cadene2019rubi}. As illustrated in Figure~\ref{fig:comp}\textcolor{red}{(c)}, given a triplet $(v,q,a)$ where $a$ is the ground-truth answer of image-question pair $(v,q)$, the branches are optimized by minimizing the cross-entropy losses over the scores $Z_{q,v,k}$, $Z_q$ and $Z_v$:
\begin{equation}\label{eq:loss}
    \mathcal{L}_{cls}=\mathcal{L}_{V\!Q\!A}(v,q,a)+\mathcal{L}_{Q\!A}(q,a)+\mathcal{L}_{V\!A}(v,a),
\end{equation}
where $\mathcal{L}_{V\!Q\!A}$, $\mathcal{L}_{Q\!A}$ and $\mathcal{L}_{V\!A}$ are over $Z_{q,v,k}$, $Z_{q}$ and $Z_{v}$. Note that we introduce a learnable parameter $c$ in Eq.~\eqref{eq:zq}-\eqref{eq:zk}, which controls the sharpness of the distribution of $Z_{q,v^*,k^*}$ like the softmax temperature~\cite{hinton2015distilling}. We hypothesize that the sharpness of NDE should be similar to that of TE. Otherwise, an improper $c$ would lead to the result that TIE in Eq.~\eqref{eq:TIE} is dominated by TE or NDE. Thus, we use Kullback-Leibler divergence to estimate $c$:
\begin{equation}
    \mathcal{L}_{kl}=\frac{1}{|\mathcal{A}|}\sum_{a\in\mathcal{A}}-p(a|q,v,k)\log p(a|q,v^*,k^*),
\end{equation}
where $p(a|q,v,k)\!=\!\softmax(Z_{q,v,k})$ and $p(a|q,v^*,k^*)\!=\!\softmax(Z_{q,v^*,k^*})$. Only $c$ is updated when minimizing $\mathcal{L}_{kl}$. 
The final loss is
the combination of $\mathcal{L}_{cls}$ and $\mathcal{L}_{kl}$:
\begin{equation}
    \mathcal{L}=\sum_{(v,q,a)\in\mathcal{D}}\mathcal{L}_{cls}+\mathcal{L}_{kl}
\end{equation}

\noindent \textbf{Inference}. As discussed in Section~\ref{sec:celook}, we use the debiased causal effect for inference, which is implemented as:
\begin{equation}\label{eq:inf}
\begin{split}
\TIE=\TE-\NDE&=Z_{q,v,k}-Z_{q,v^*,k^*}\\
&=h(z_q,z_v,z_k)-h(z_q,z^*_v,z^*_k).
\end{split}
\end{equation}
\begin{table*}[t]
\centering
\caption{\textbf{Comparison on VQA-CP v2 test set and VQA v2 val set}. \underline{\bf Best} and \textbf{second best} results obtained without extra generated training samples are highlighted in each column. ``Base.'' indicates the VQA base model. We report the average accuracy of CF-VQA over 5 experiments with different random seeds.}
\label{tab:all}
\vspace{-2mm}
\scalebox{0.81}{
\begin{tabular}{l c c c cccc c cccc}
\hline
\toprule
Test set & & & & \multicolumn{4}{c}{VQA-CP v2 test} & & \multicolumn{4}{c}{VQA v2 val}\\  
\cmidrule{5-8} \cmidrule{10-13}
Methods & ~ & Base. & ~ & {All} & {Y/N} &  {Num.} & {Other} & ~ & {All} & {Y/N} & {Num.} & {Other}\\
\midrule
GVQA~\cite{agrawal2018don}   &  & -- & &  31.30 & 57.99 & 13.68  & 22.14 & & 48.24 & 72.03   & 31.17  & 34.65 \\
SAN~\cite{yang2016stacked}  &  & -- & &   24.96 & 38.35 & 11.14  & 21.74 & & 52.41 & 70.06 & 39.28 & 47.84 \\
UpDn~\cite{anderson2018bottom}  &  & -- & & 39.74 & 42.27 & 11.93 & 46.05 & & 63.48 & 81.18 & 42.14 & \underline{\bf 55.66} \\
S-MRL~\cite{cadene2019rubi} & & -- & &  38.46 & 42.85 & 12.81 & 43.20 & & 63.10 & -- & -- & -- \\

\hline
\multicolumn{13}{l}{\it methods based on  modifying language module follow:} \\
\hline
DLR~\cite{jing2020overcoming} &  & UpDn &  &  48.87 & 70.99 & 18.72 & 45.57  & & 57.96 & 76.82 & 39.33 & 48.54 \\
VGQE~\cite{gouthaman2020reducing} &  & UpDn &  & 48.75 & -- & -- & -- & & \underline{\bf 64.04} & -- & -- & -- \\
VGQE~\cite{gouthaman2020reducing} &  & S-MRL &  &  50.11 & 66.35 & \bf 27.08 & 46.77 & & 63.18 & -- & -- & -- \\

\hline
\multicolumn{13}{l}{\it methods based on strengthening visual attention follow:} \\
\hline
AttAlign \cite{selvaraju2019taking} & & UpDn & &  39.37 & 43.02 &11.89 & 45.00 & & 63.24 & 80.99 & 42.55 & 55.22 \\
HINT   \cite{selvaraju2019taking} & & UpDn & &  46.73 & 67.27 & 10.61 & 45.88 & & 63.38 & 81.18 & 42.99 & \bf 55.56 \\
SCR~\cite{wu2019self} & & UpDn & & 49.45 & 72.36 & 10.93 & \underline{\bf 48.02} & & 62.2 & 78.8 & 41.6 & 54.5 \\

\hline
\multicolumn{13}{l}{\it methods based on weakening language prior follow:} \\
\hline
AdvReg.    \cite{ramakrishnan2018overcoming} & & UpDn & &  41.17 & 65.49 & 15.48 & 35.48 & & 62.75 & 79.84 & 42.35 & 55.16 \\
RUBi~\cite{cadene2019rubi} & & UpDn & & 44.23 & 67.05 & 17.48 & 39.61 & & --  &  --  & --  & -- \\
RUBi~\cite{cadene2019rubi} & & S-MRL & &  47.11 & 68.65 & 20.28 & 43.18 & & 61.16 &  --  & --  & -- \\
LM~\cite{clark2019don} & & UpDn & & 48.78 & 72.78 & 14.61 & 45.58 & & 63.26 & 81.16 & 42.22 & 55.22 \\
LM+H~\cite{clark2019don} & & UpDn & & 52.01 & 72.58 & \underline{\bf 31.12} & \bf 46.97 & &56.35 & 65.06 & 37.63 & 54.69 \\
 
\midrule

CF-VQA (HM) & & UpDn & & 49.74$^{\pm 0.26}$ & 74.81$^{\pm 0.51}$ & 18.46$^{\pm 0.39}$ & 45.19$^{\pm 0.27}$ &  & \bf 63.73$^{\pm 0.07}$ & \bf 82.15$^{\pm 0.08}$ & \bf 44.29$^{\pm 0.54}$ & 54.86$^{\pm 0.07}$ \\

CF-VQA (HM) & & S-MRL & & 51.27$^{\pm 0.25}$ & 77.80$^{\pm 0.83}$ & 20.64$^{\pm 1.01}$ & 45.76$^{\pm 0.21}$ & & 62.49$^{\pm 0.06}$ & 81.19$^{\pm 0.09}$ & \underline{\bf 44.64}$^{\pm 0.20}$ & 52.98$^{\pm 0.14}$ \\
CF-VQA (SUM) & & UpDn & & \bf 53.55$^{\pm 0.10}$ & \underline{\bf 91.15}$^{\pm 0.06}$ & 13.03$^{\pm 0.21}$ & 44.97$^{\pm 0.20}$ & & 63.54$^{\pm 0.09}$ & \underline{\bf 82.51}$^{\pm 0.12}$ & 43.96$^{\pm 0.17}$ & 54.30$^{\pm 0.09}$ \\

CF-VQA (SUM) & & S-MRL & & \underline{\bf 55.05}$^{\pm 0.12}$ & \bf 90.61$^{\pm 0.28}$ & 21.50$^{\pm 0.86}$ & 45.61$^{\pm 0.16}$ & & 60.94$^{\pm 0.15}$ & 81.13$^{\pm 0.15}$ & 43.86$^{\pm 0.40}$ & 50.11$^{\pm 0.16}$ \\

\midrule
\hline
\multicolumn{13}{l}{\it methods based on balancing training data follow:} \\
\hline
CVL~\cite{abbasnejad2020counterfactual} &  & UpDn &  &  42.12 & 45.72 & 12.45 & 48.34  & & -- & -- & -- & -- \\
Unshuffling~\cite{teney2020unshuffling} &  & UpDn &  &  42.39 & 47.72 & 14.43 & 47.24  & & 61.08 & 78.32 & 42.16 & 52.81 \\
RandImg~\cite{teney2020value} &  & UpDn &  &  55.37 & 83.89 & 41.60 & 44.20 & & 57.24 & 76.53 & 33.87 & 48.57 \\
SSL~\cite{zhu2020overcoming} & & UpDn &  & 57.59 & 86.53 & 29.87 & 50.03
 & & 63.73 & -- & -- & -- \\
CSS~\cite{chen2020counterfactual} & & UpDn &  & 58.95 & 84.37 & 49.42 & 48.21
 & & 59.91 & 73.25 & 39.77 & 55.11 \\
CSS+CL~\cite{liang2020learning} & & UpDn &  & 59.18 & 86.99 & 49.89 & 47.16 & & 57.29 & 67.27 & 38.40 & 54.71 \\
Mutant~\cite{gokhale2020mutant} & & UpDn &  &  61.72
 & 88.90 & 49.68 & 50.78
 & & 62.56 & 82.07 & 42.52 & 53.28 \\
\bottomrule
\end{tabular}
}
\end{table*}

\subsection{Revisiting RUBi and Learned-Mixin}\label{sec:revisit}

\begin{figure}
\centering
{
\subfigure[]{
\label{fig:vqacg-simple-1}
\includegraphics[width=0.3\linewidth]{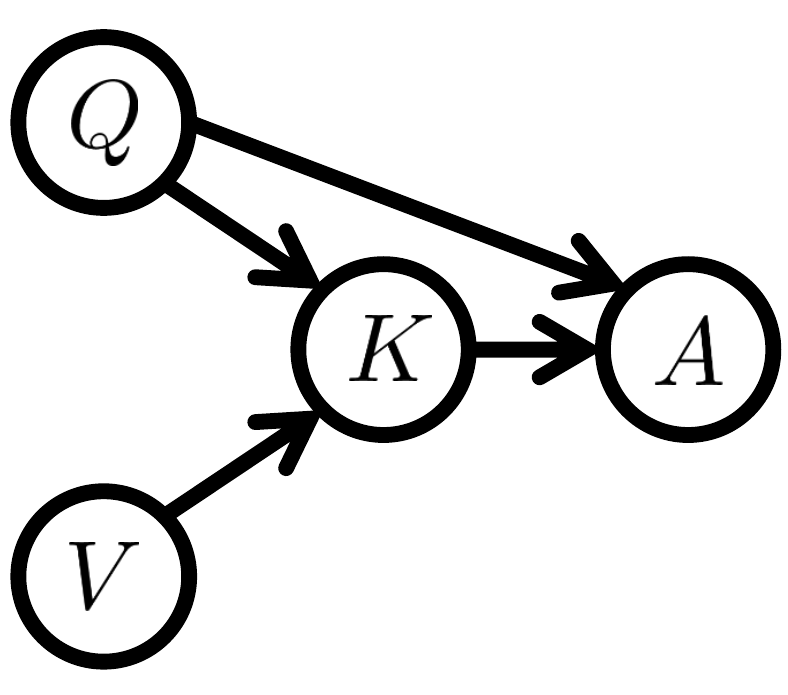}
}
\subfigure[]{
\label{fig:vqacg-simple-2}
\includegraphics[width=0.64\linewidth]{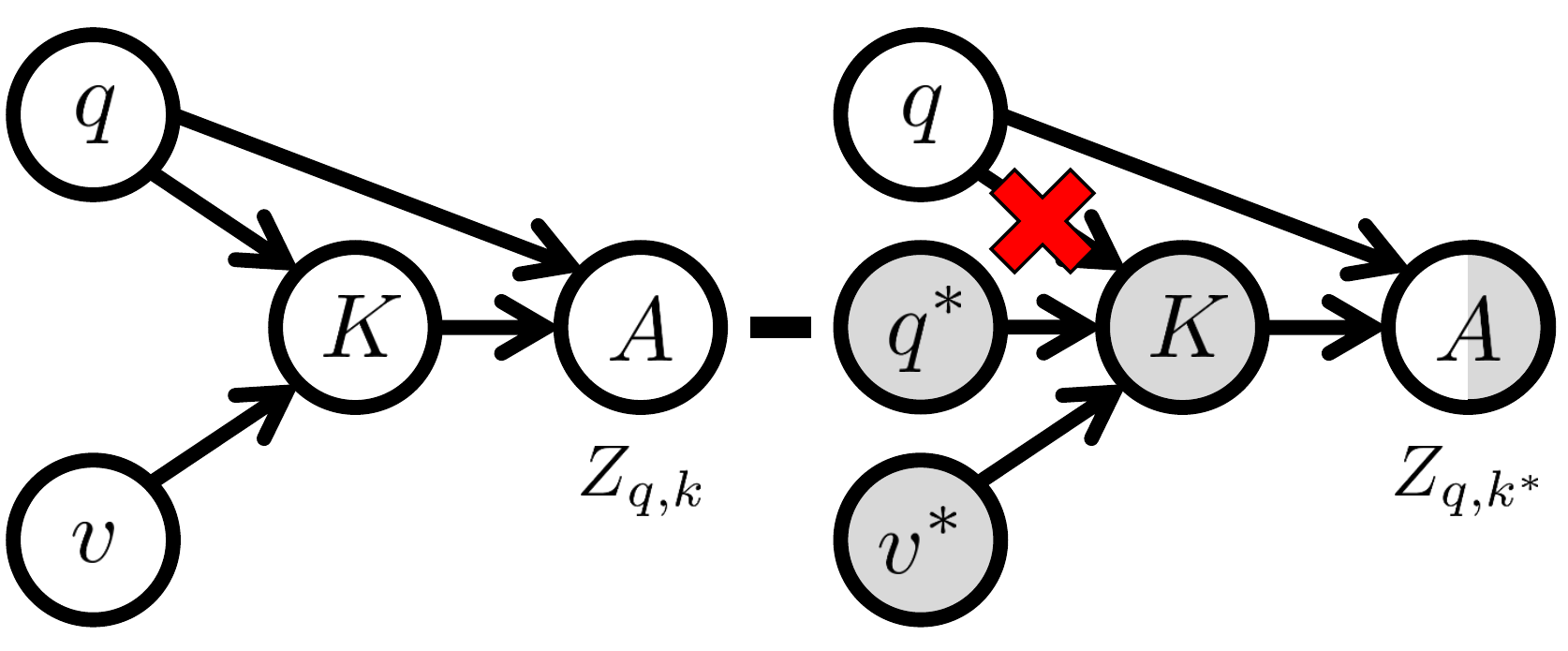}
}
}
\vspace{-2mm}
\caption{(a) Simplified VQA causal graph. (b) Comparison between conventional VQA and counterfactual VQA.}
\label{fig:vqacg-simple}
\vspace{-2mm}
\end{figure}

Note that recent language-prior based methods RUBi~\cite{cadene2019rubi} and Learned-Mixin~\cite{clark2019don} use an ensemble model that consists a vision-language branch $\mathcal{F}_{VQ}$ and a question-only branch $\mathcal{F}_{Q}$. Besides, they simply exclude $\mathcal{F}_{Q}$ and use $z_k\!=\!\mathcal{F}_{VQ}(v,q)$ during the inference stage. The conceptual comparison between our CF-VQA and language-prior based strategies are illustrated in Figure~\ref{fig:comp}. These methods can be unified into our counterfactual inference framework, which (1) follow a simplified causal graph 
(Fig.~\ref{fig:vqacg-simple-1}) 
without the direct path $V\!\rightarrow\!A$, and (2) use natural indirect effect (NIE) in Eq.~\eqref{eq:NIE} for inference. The detailed analysis is provided in the appendix.
\section{Experiments}\label{sec:exp}
We mainly conduct the experiments on the VQA-CP~\cite{agrawal2018don} dataset. VQA-CP is proposed to evaluate the robustness of VQA models when the answer distributions of training and test splits are significantly different. In addition, we also report the results on the balanced VQA v2 dataset to see whether the approach over-corrects language bias. The models are evaluated via accuracy. 
We conduct experiments with three baseline VQA architectures: Stacked Attention Network (\textbf{SAN})~\cite{yang2016stacked}, Bottom-up and Top-down Attention (\textbf{UpDn})~\cite{anderson2018bottom}, and a simplified MUREL~\cite{cadene2019murel} (\textbf{S-MRL})~\cite{cadene2019rubi}.

\begin{table*}[t]
\centering
\caption{\textbf{Ablation of CF-VQA} on VQA-CP v2 test set. ``SAN/UpDn/S-MRL'' denotes the baseline VQA model. ``HM/SUM'' represents the strategies that train the ensemble model and test with only the vision-language branch following ensemble-based method~\cite{cadene2019rubi,clark2019don}. $^*$ represents the reproduced results.}
\label{tab:1-1-ablation}
\centering
\vspace{-2mm}
\scalebox{0.81}{
    \begin{tabular}{l cccc}
        \hline
        \toprule
          & All & {Y/N} &  {Num.} & {Other}\\
        \hline
        SAN$^*$ & 33.18 & 38.57 & 12.25 & 36.10 \\
        \hline
        HM & 45.89 & 70.37 & \bf 23.99 & 39.07 \\
        +~CF-VQA & \bf 48.10 & \bf 77.68 & 22.19 & \bf 39.71 \\
        \hline
        SUM & 43.98 & 68.98 & \bf 17.32 & 38.19\\
        +~CF-VQA & \bf 50.15 & \bf 87.95 & 16.46 & \bf 39.59 \\
        \bottomrule
        \hline
    \end{tabular}
}
\hfill
\scalebox{0.81}{
    \begin{tabular}{l cccc}
        \hline
        \toprule
          & All & {Y/N} &  {Num.} & {Other}\\
        \hline
        UpDn$^*$ & 37.69 & 43.17 & 12.53 & 41.72 \\
        \hline
        HM & 47.97 & 69.19 & \bf 18.80 & 44.86 \\
        +~CF-VQA & \bf 49.74 & \bf 74.81 & 18.46 & \bf 45.19 \\
        \hline
        SUM & 47.29 & 72.26 & 12.54 & 43.74 \\
        +~CF-VQA & \bf 53.55 & \bf 91.15 & \bf 13.03 & \bf 44.97 \\
        \bottomrule
        \hline
    \end{tabular}
}
\hfill
\scalebox{0.81}{
    \begin{tabular}{l cccc}
        \hline
        \toprule
          & All & {Y/N} &  {Num.} & {Other} \\
        \hline
        S-MRL$^*$ & 37.09 & 41.39 & 12.46 & 41.60 \\
        \hline
        HM & 49.37 & 73.20 & 20.10 & 44.92 \\
        +~CF-VQA & \bf 51.27 & \bf 77.80 & \bf 20.64 & \bf 45.76 \\
        \hline
        SUM & 48.27 & 74.60 & 20.96 & 41.96 \\
        +~CF-VQA & \bf 55.05 & \bf 90.61 & \bf 21.50 & \bf 45.61 \\
        \bottomrule
        \hline
    \end{tabular}
}
\vspace{-2mm}
\end{table*}

\begin{table*}[t]
\centering
\caption{\textbf{Ablation of CF-VQA with the simplified causal graph} on VQA-CP v2 test set. ``SAN/UpDn/S-MRL'' denotes the baseline VQA model. ``HM/SUM'' represents the strategies that train the ensemble model and test with only the vision-language branch following ensemble-based method~\cite{cadene2019rubi,clark2019don}. $^*$ represents the reproduced results.}
\label{tab:1-2-ablation-sim}
\centering
\vspace{-2mm}
\scalebox{0.81}{
    \begin{tabular}{l cccc}
        \hline
        \toprule
          & All & {Y/N} &  {Num.} & {Other}\\
        \hline
        SAN$^*$ & 33.18 & 38.57 & 12.25 & 36.10 \\
        \hline
        HM & 45.48 & 71.32 & 17.19 & 39.71 \\
        +~CF-VQA & \bf 49.43 & \bf 83.82 & \bf 17.52 & \bf 40.16 \\
        \hline
        SUM & 42.35 & 62.33 & \bf 16.64 & 38.94 \\
        +~CF-VQA & \bf 49.85 & \bf 87.75 & 16.15 & \bf 39.24 \\
        \bottomrule
        \hline
    \end{tabular}
}
\hfill
\scalebox{0.81}{
    \begin{tabular}{l cccc}
        \hline
        \toprule
          & All & {Y/N} &  {Num.} & {Other}\\
        \hline
        UpDn$^*$ & 37.69 & 43.17 & 12.53 & 41.72 \\
        \hline
        HM & 46.50 & 67.54 & 12.83 & 44.72 \\
        +~CF-VQA & \bf 49.53 & \bf 77.02 & \bf 12.86 & \bf 45.18 \\
        \hline
        SUM & 47.10 & 70.00 & 12.80 & 44.51 \\
        +~CF-VQA & \bf 53.55 & \bf 91.15 & \bf 12.81 & \bf 45.02 \\
        \bottomrule
        \hline
    \end{tabular}
}
\hfill
\scalebox{0.81}{
    \begin{tabular}{l cccc}
        \hline
        \toprule
          & All & {Y/N} &  {Num.} & {Other} \\
        \hline
        S-MRL$^*$ & 37.09 & 41.39 & 12.46 & 41.60 \\
        \hline
        HM & 49.57 & 72.31 & 20.28 & 45.68 \\
        +~CF-VQA & \bf 52.68 & \bf 82.05 & \bf 20.76 & \bf 46.04 \\
        \hline
        SUM & 49.42 & 74.43 & 20.52 & 44.24 \\
        +~CF-VQA & \bf 54.52 & \bf 90.69 & \bf 21.84 & \bf 44.53 \\
        \bottomrule
        \hline
    \end{tabular}
}
\vspace{-4mm}
\end{table*}

\subsection{Quantitative Results}
We first compare CF-VQA with state-of-the-art methods. Recent approaches can be grouped as follows. (1) Methods that \bfit{modify language modules} proposed to decouple the linguistic concepts (\textbf{DLR})~\cite{jing2020overcoming} or generate visually-grounded question representations (\textbf{VGQE})~\cite{gouthaman2020reducing}. (2) Methods that \bfit{strengthen visual attention} exploit human visual~\cite{das2017human} or textual~\cite{huk2018multimodal} explanations, including \textbf{AttAlign}~\cite{selvaraju2019taking}, \textbf{HINT}~\cite{selvaraju2019taking} and \textbf{SCR}~\cite{wu2019self}.
(3) Methods that \bfit{weaken language prior} proposed to directly formulate the language prior by a separate question-only branch, including \textbf{AdvReg.}~\cite{ramakrishnan2018overcoming}, \textbf{RUBi}~\cite{cadene2019rubi} and \textbf{Learned-Mixin (LM)}~\cite{clark2019don}. 
(4) Methods that \bfit{balance training data} proposed to change the training distribution for unbiased training, including \textbf{CVL}~\cite{abbasnejad2020counterfactual}, \textbf{Unshuffling}~\cite{teney2020unshuffling}, \textbf{RandImg}~\cite{teney2020value}, \textbf{SSL}~\cite{zhu2020overcoming}, \textbf{CSS}~\cite{chen2020counterfactual}, \textbf{CL}~\cite{liang2020learning}, and \textbf{Mutant}~\cite{gokhale2020mutant}. 
Unshuffling~\cite{teney2020unshuffling} partitions the training set into multiple invariant subsets. Other methods generate counterfactual training samples by masking or transforming critical words and objects~\cite{chen2020counterfactual,gokhale2020mutant} or replacing the image~\cite{abbasnejad2020counterfactual,teney2020value,zhu2020overcoming}.

\begin{table}
\centering
\caption{\textbf{Ablation of assumptions for counterfactual outputs} on VQA-CP v2 test set.}
\label{tab:3-prior}
\vspace{-1mm}
\scalebox{0.85}{
\begin{tabular}{l l cccc}
    \hline
    \toprule
      & & All & {Y/N} &  {Num.} & {Other}\\
    \midrule
    S-MRL~\cite{cadene2019rubi}  &  & 38.46 & 42.85 & 12.81 & 43.20 \\
    \hline
    \multirow{3}{*}{HM} & random & 31.27 & 29.69 & \bf 42.87 & 28.91 \\
    & prior & 46.29 & 61.88 & 20.03 & 45.33 \\
    & uniform & \bf 51.27 & \bf 77.80 & 20.64 & \bf 45.76 \\
    
    \midrule
    \multirow{3}{*}{SUM} & random & 27.52 & 28.00 & \bf 37.88 & 24.42 \\
     & prior & 38.06 & 41.43 & 14.90 & 42.64 \\
     & uniform & \bf 55.05 & \bf 90.61 & 21.50 & \bf 45.61 \\
    \bottomrule
    \hline
\end{tabular}
}
\vspace{-4mm}
\end{table}

The results on VQA-CP v2 and VQA v2 are reported in Table~\ref{tab:all}. Most of the methods that explicitly generate training samples~\cite{teney2020value,zhu2020overcoming,chen2020counterfactual,liang2020learning,gokhale2020mutant} outperform others by large margins. However, these methods explicitly change the training priors, which violates the original intention of VQA-CP, \ie, to evaluate whether VQA models are driven by memorizing priors in training data~\cite{agrawal2018don}. Therefore, we do not directly compare CF-VQA with these methods for fairness. Overall, compared to non-augmentation approaches, our proposed CF-VQA achieves a new state-of-the-art performance on VQA-CP v2. With a deep look at the question type, we find that the improvement on ``Yes/No'' questions is extremely large (from $\sim$70\% to $\sim$90\%), which indicates that language bias have different effects on different types of questions. Besides, methods with extra annotations or generated training samples effectively improves the accuracy on ``Other'' questions, while CF-VQA achieves comparable performance to other methods. It is worth noting that LM~\cite{clark2019don} achieves a competitive performance on VQA-CP v2 with an additional language entropy penalty (LM+H). However, the accuracy drops significantly by $\sim$7\% on VQA v2, which indicates that the entropy penalty forces the model to over-correct language bias, especially on ``Yes/No'' questions. As a comparison, CF-VQA is more robust on VQA v2.

\begin{table}[t]
\centering
\caption{The comparison between CF-VQA and RUBi.}
\label{tab:5-rubi}
\vspace{-.5em}
\centering
\scalebox{0.8}{
\begin{tabular}{l cccc c c}
    \hline
    \toprule
      & \multicolumn{4}{c}{VQA-CP v2 test} & & VQA v2 val\\
    \cmidrule{2-5} \cmidrule{7-7}
      & All & {Y/N} &  {Num.} & {Other} &  & All \\
    \hline
    S-MRL~\cite{cadene2019rubi}  & 38.46 & 42.85 & 12.81 & \bf 43.20 &  & \bf 63.10 \\
    \hline
    RUBi~\cite{cadene2019rubi}  & 47.11$^{\pm 0.51}$ & 68.65 & 20.28 & 43.18 &  & 61.16\\
    +~CF-VQA & \bf 54.69$^{\pm 0.98}$ & \bf 89.90 & \bf 32.39 & 42.01 &  & 60.53 \\
    \bottomrule
    \hline
\end{tabular}
}
\vspace{-4mm}
\end{table}

\begin{figure*}
\centering
\includegraphics[width=\textwidth]{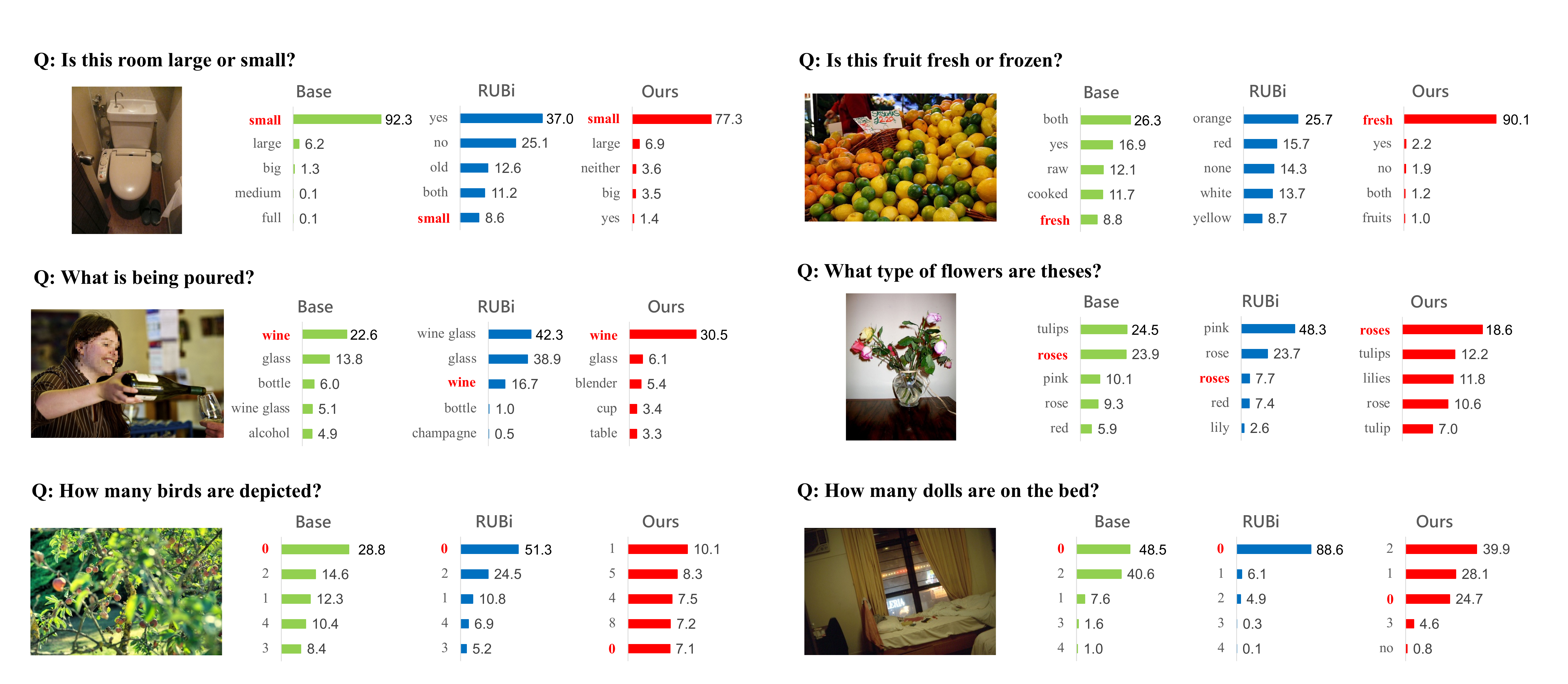}
\caption{\textbf{Qualitative comparison} on VQA-CP v2 test split. Red bold answer denotes the ground-truth one.}
\label{fig:ex}
\vspace{-3mm}
\end{figure*}

\begin{figure}
\centering
\includegraphics[width=\linewidth]{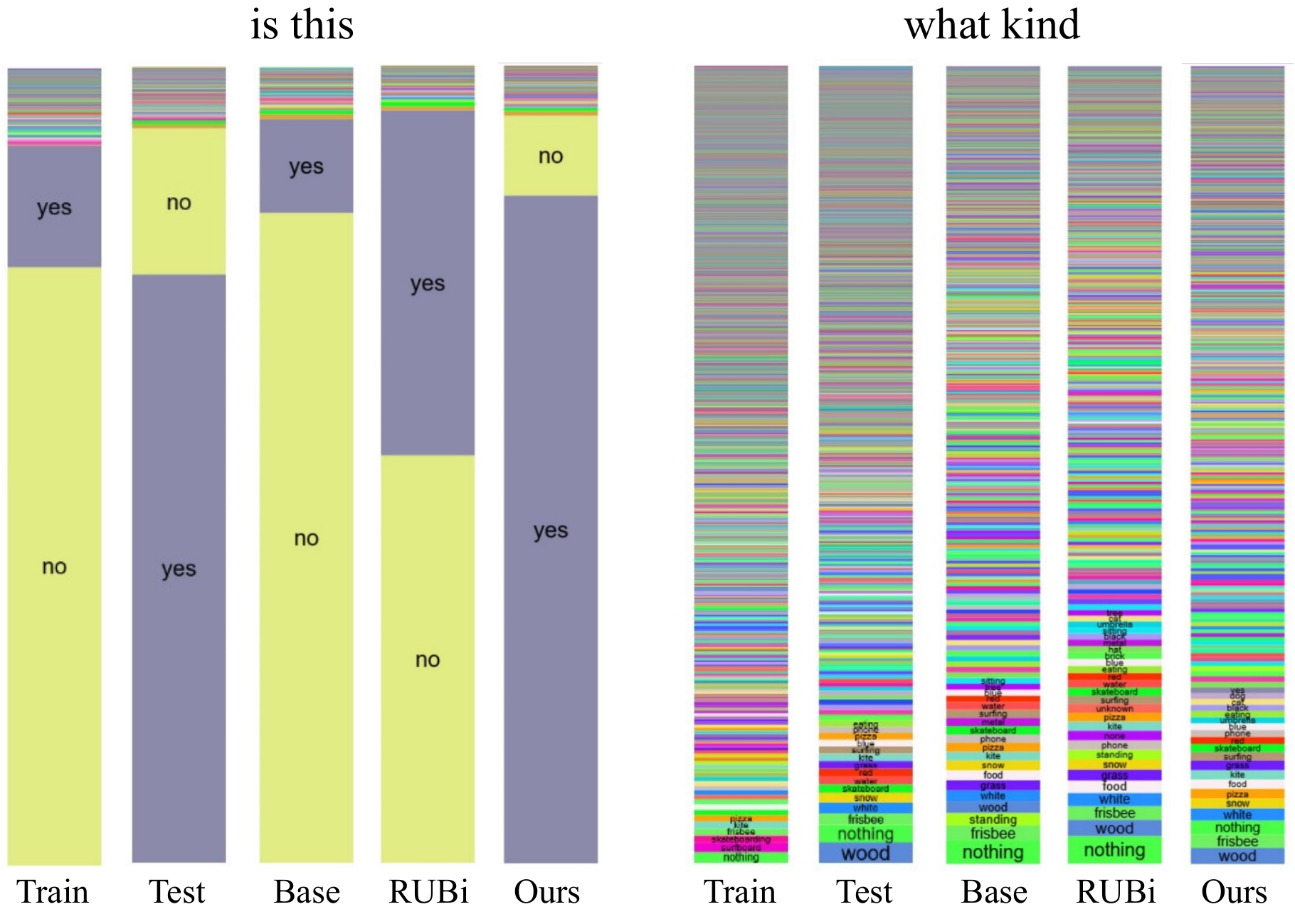}
\caption{\textbf{Answer distributions} on VQA-CP v2.}
\label{fig:distribution}
\vspace{-4mm}
\end{figure}

We further conduct ablation studies to validate (1) the generalizability of CF-VQA to both baseline VQA architectures, fusion strategies and causal graphs, and (2) the distribution assumption under the no-treatment conditions. As shown in Table~\ref{tab:1-1-ablation} and~\ref{tab:1-2-ablation-sim}, CF-VQA outperforms ensemble-based strategies by over 2\% for HM and over 5\% for SUM in all cases. To empirically validate the distribution assumption, we proposed two candidate assumptions. ``Random'' denotes that $c_i$ for answer $a_i$ are learned without any constraint. ``Prior'' denotes that $\{c_i\}$ obey the prior distribution of the training set. As shown in Table~\ref{tab:3-prior}, ``random'' and ``prior'' even perform worse than the baselines. As we discussed in Section~\ref{sec:imp}, the possible reason is that the uniform distribution assumption guarantees a safe estimation of NDE, \ie, the biased language effect.

Based on our cause-effect look, CF-VQA can easily improve RUBi by replacing NIE with TIE and introducing only one learnable parameter. The details are provided in the appendix. As shown in Table~\ref{tab:5-rubi}, CF-VQA improves RUBi by over 7.5\% (\ie, from 47.11 to 54.69) on VQA-CP v2, while the accuracy slightly drops on VQA v2. Compared to our proposed symmetric fusion strategies HM and SUM, the standard deviation of RUBi's accuracy is larger, which indicates that our proposed symmetric fusion strategies are more stable and robust.

\subsection{Qualitative Results}

The qualitative results are provided to validate whether CF-VQA can effectively reduce language bias and retain language context. 
As illustrated in Figure~\ref{fig:distribution}, CF-VQA can successfully overcome language bias on yes/no questions compared to RUBi, while the baseline model suffers from the memorized language prior on the training set. Besides, for ``what kind'' questions, RUBi prefers the meaningless answer ``none'' rather than specific ones. Although CF-VQA cannot recover the answer distribution very well, it attempts to respond with more meaningful answers (\eg, wood, frisbee).
Examples in Figure~\ref{fig:ex} further illustrate how CF-VQA preserves language context for inference. For the left top example, CF-VQA recognizes the correct context ``large or small'', while RUBi tends to answer yes/no based on the wrong context ``is this''. For the right example at the second row, although RUBi successfully locates the flowers, it wrongly focuses on visual attributes (\ie, ``pink'') rather than categories (\ie, ``what type''). These results highlight the importance of language context, which is not considered by language prior based approaches. The examples at the third row show the failure cases on number-related questions. It remains a open problem how to improve the accuracy of number-related questions. 
\section{Conclusion}
In this paper, we proposed a novel counterfactual inference framework CF-VQA to reduce language bias in VQA. The bias is formulated as the direct causal effect of questions on answers and estimated by counterfactual reasoning. The reduction of language bias is realized by subtract the direct linguistic effect from the total causal effect. Experimental results demonstrate the effectiveness and generalizability of CF-VQA. Furthermore, recent debiasing studies~\cite{cadene2019rubi,clark2019don} can be unified into our proposed counterfactual inference framework. Surprisingly, we can further improve RUBi~\cite{cadene2019rubi} by simply changing a few lines of code and including only one more learnable parameter based on our cause-effect look. In the future, we will consider how to make balances between robustness and debiasing ability.

\noindent\textbf{Acknowledgements}~~~We would like to thank anonymous ACs and reviewers for their valuable discussion and insightful suggestions.
This work was supported in part by NTU-Alibaba JRI, MOE AcRF Tier 2 grant, National Natural Science Foundation of China (61976220 and 61832017), and Beijing Outstanding Young Scientist Program (BJJWZYJH012019100020098). 

This supplementary document is organized as follows:
\begin{itemize}
\item Section~\ref{sec:supp-1} introduces that RUBi~\cite{cadene2019rubi} and Learned-Mixin~\cite{clark2019don} can be unified into our counterfactual inference framework.

\item Section~\ref{sec:supp-2} provides an analysis of estimating NDE using the learnable parameter.

\item Section~\ref{sec:supp-3} describes the implementation details.

\item Section~\ref{sec:supp-4} describes the supplementary quantitative and qualitative results.
\end{itemize}

\section{Revisiting RUBi and Learned-Mixin}\label{sec:supp-1}

As mentioned in Section 4.3, RUBi~\cite{cadene2019rubi} and Learned-Mixin~\cite{clark2019don} can be unified into our counterfactual inference framework, which (1) follow a simplified causal graph without the direct path $V\!\rightarrow\!A$, and (2) use natural indirect effect (NIE) for inference. The detailed analysis is provided as follows.

\subsection{Cause-Effect Look}
\begin{figure*}
    \centering
    \includegraphics[width=.88\linewidth]{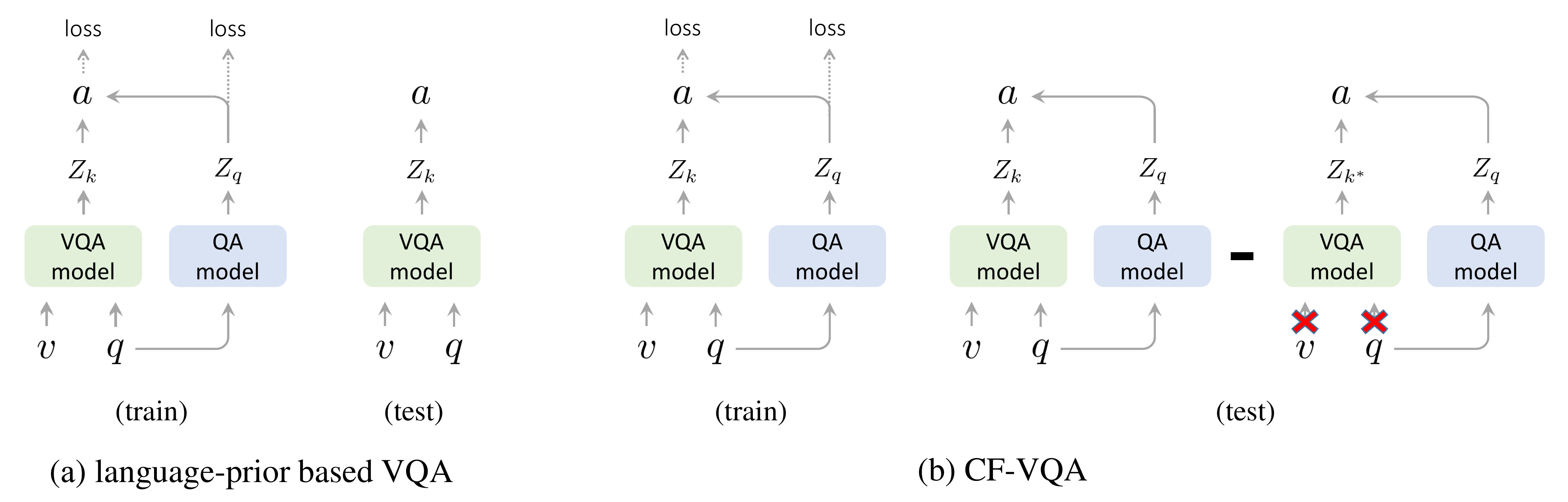}
    \caption{Comparison between our CF-VQA and language-prior based methods~\cite{cadene2019rubi,clark2019don} based on the simplified causal graph. }
    \label{fig:supp-comp}
\end{figure*}

Recent works RUBi~\cite{cadene2019rubi} and Learned-Mixin~\cite{clark2019don} apply an ensemble architecture with a vision-language branch $\mathcal{F_{VQ}}$ and a question-only branch $\mathcal{F_{Q}}$, while the direct relation between vision and answer is not formulated. The architecture is shown in Figure~\ref{fig:supp-comp} \textcolor{red}{(a)}.

Note that total effect can be decomposed into natural direct effect (NDE) and total indirect effect (TIE). As introduced in the main paper, we remove language bias by subtracting the natural direct effect from the total effect. The TIE is calculated by:
\begin{equation}\label{eq:supp-vqatie}
\begin{split}
    \TE &= Z_{q,k}-Z_{q^*,k^*},\\
    \NDE &= Z_{q,k^*}-Z_{q^*,k^*},\\
    \TIE &= \TE-\NDE = Z_{q,k}-Z_{q,k^*},
\end{split}
\end{equation}
which corresponds to Eq. (4) in the main paper. An alternative option to reduce language bias is to substract the total direct effect (TDE) of questions on answers from total effect, which is formulated as:
\begin{equation}\label{eq:supp-vqanie}
\begin{split}
    \TDE &= Z_{q,k}-Z_{q^*,k},\\
    \NIE &= \TE-\TDE = Z_{q^*,k}-Z_{q^*,k^*}.
\end{split}
\end{equation}
Intuitively, both TIE and NIE reflect the increase of confidence for the answer given the visual knowledge, \ie, from $k^*$ to $k$. The difference between TIE and NIE is the existence of question $q$. The question $q$ is block to calculate NIE (\ie, $q^*$), while $q$ is given to calculate TIE. We use TIE to reserve $q$ as the language context. In addition, both TDE and NDE reflect the increase of confidence for the answer given the question, \ie, from $q^*$ to $q$. The difference between TDE and NDE is also the existence of question $q$. Note that we hope to exclude the effect directly caused by question. Therefore, the mediator knowledge should be blocked when estimating the pure language effect, which is captured by NDE.

\subsection{Implementation}
RUBi~\cite{cadene2019rubi} and Learned-Mixin (LM)~\cite{clark2019don} use the following fusion strategies for ensemble-based training:
\begin{flalign}\label{eq:supp-fusion-rubi}
\text{(RUBi)}&&h(Z_q,Z_k)=Z_k\cdot\sigma(Z_q)&&\phantom{\text{(RUBi)}}
\end{flalign}
\begin{flalign}\label{eq:supp-fusion-lm}
\text{(LM)}&&h(Z_q,Z_k)=\log\sigma(Z_k)+g(k)\cdot\log\sigma(Z_q)&&\phantom{\text{(LM)}}
\end{flalign}
where $\sigma(\cdot)$ represents the sigmoid function, and $g(\cdot)$ is a learned function $\mathbb{R}^d\rightarrow \mathbb{R}^1$ with the knowledge representation $k\in\mathbb{R}^d$ as input and a scalar weight as output. During the test stage, they use $Z_k$ for inference.

\begin{algorithm}[t]
    \centering
    \caption{Improving RUBi~\cite{cadene2019rubi} using CF-VQA}\label{alg:supp-rubi}
    \begin{algorithmic}[1]
    \Function {RUBi}{$v, ~q, ~\text{is\_Training};~\theta, ~\textcolor{red}{c}$}
        \State $z_q=\mathcal{F}_Q(q)$
        \State $z_k=\mathcal{F}_{VQ}(v,~q)$
        \If {\text{is\_Training}} 
            \State $z=z_k\cdot\sigma(z_q)$
            \State updating $\theta$ according to $\mathcal{L}_{cls}$
            \State 
            \textcolor{red}{updating $c$ according to $\mathcal{L}_{kl}$}
        \Else
            \State \st{$z=z_{k}$}~\textcolor{red}{ $z=(z_k-c)\cdot\sigma(z_q)$}
        \EndIf
        \State \textbf{return} $z$
    \EndFunction
    \end{algorithmic}
\end{algorithm}

Perhaps supering
As for RUBi, NIE is calculated as:
\begin{equation}
    \NIE=\underbrace{z_k\cdot\sigma(c)}_{Z_{q^*,k}}-\underbrace{c\cdot\sigma(c)}_{Z_{q^*,k^*}}\propto z_k
\end{equation}
As for Learned-Mixin, NIE is calculated as:
\begin{equation}
\begin{split}
    \NIE=&\underbrace{\left(\log\sigma(z_k)+g(k)\cdot\log\sigma(c)\right)}_{Z_{q^*,k}}\\
    &-\underbrace{\left(\log\sigma(c)+g(k^*)\cdot\log\sigma(c)\right)}_{Z_{q^*,k^*}}\propto z_k\\
\end{split}
\end{equation}
where $c$, $g(k)$ and $g(k^*)$ are constants for the same sample. Therefore, we have $\NIE\!\propto\!z_k$ for both RUBi and Learned-Mixin, which is exactly the output score of the vision-language branch $\mathcal{F}_{VQ}$. Note that RUBi and Learned-Mixin simply preserve the vision-language branch and uses $z_k$ for inference. From our cause-effect view, \textit{RUBi and Learned-Mixin use natural indirect effect for inference}.

\subsection{Improving RUBi~\cite{cadene2019rubi}}\label{sec:supp-imp}
Thanks to our cause-effect look, RUBi~\cite{cadene2019rubi} can be improved using CF-VQA, \ie, using TIE for inference. Specifically, TIE for RUBi is calculated as:
\begin{equation}\label{eq:supp-tie-rubi}
\TIE=\underbrace{z_k\cdot\sigma(z_q)}_{Z_{q,k}}-\underbrace{c\cdot\sigma(z_q)}_{Z_{q,k^*}}
\end{equation}
where $c$ denotes a learnable parameter. Table 5 in the main paper demonstrates that CF-VQA can outperform RUBi by 7\% on VQA-CP v2. The red notes in Algorithm~\ref{alg:supp-rubi} show how RUBi is improved by changing several lines of code.

\begin{table*}[t]
\centering
\caption{\textbf{Comparison on VQA-CP v2 val set}. ``Base.'' indicates the VQA base model.}
\label{tab:supp-1-val}
\vspace{-.5em}
\centering
\scalebox{0.9}{
\begin{tabular}{l c cccc c c c c}
    \hline
    \toprule
      & & \multicolumn{6}{c}{VQA-CP v2} & & VQA v2\\
    \cmidrule{3-8} \cmidrule{10-10}
      & & \multicolumn{4}{c}{val (in-domain)}  & & test (OOD) & & val (in-domain)\\
      \hline
      & Base. & All & {Y/N} &  {Num.} & {Other} & & All & & All \\
    \hline
    GRLS~\cite{grand2019adversarial}  & -- & 56.90 & 69.23 & 42.50 & 49.36 & & 42.33 & & 51.92 \\
    GradSup\cite{teney2020learning}  & -- & 62.4 & 77.8 & 43.8 & 53.6 & & 46.8 & & -- \\
    RandImg~\cite{teney2020value}  & UpDn & 54.24 & 64.22 & 34.40 & 50.46 & & 55.37 & & 57.24 \\
    \hline
    CF-VQA (HM) & UpDn & 65.47 & 79.09 & 45.86 & 57.86 &  & 49.74 & & 63.73 \\
    CF-VQA (SUM) & UpDn & 60.29 & 66.32 & 47.48 & 57.96 &  & 51.27 & & 62.49 \\
    CF-VQA (HM) & S-MRL & 63.08 & 75.76 & 44.88 & 55.99 &  & 53.55 & & 63.54 \\
    CF-VQA (SUM) & S-MRL & 57.86 & 66.24 & 44.98 & 53.38 &  & 55.05 & & 60.94 \\
    \bottomrule
    \hline
\end{tabular}
}
\end{table*}

\section{Analysis of Estimating NDE}\label{sec:supp-2}

In Section 4.2 in the main paper, we claimed that the learnable parameter $c$ controls the sharpness of $Z_{q,v^*,k^*}$ for estimating NDE. We give an intuitive analysis here.

For Harmonic (HM), we have:
\begin{flalign}\label{eq:supp-z-har}
\text{(HM)}&&Z_{q,v^*,k^*}=\log\frac{\sigma(z_q)\cdot c_{\text{HM}}}{1+\sigma(z_q)\cdot c_{\text{HM}}},&&
\end{flalign}
where $c_{\text{HM}}\!=\!(\sigma(c))^2\in(0,1)$. We approximate the limits of $Z_{q,v^*,k^*}$ and $\TIE=Z_{q,v,k}-Z_{q,v^*,k^*}$ as:
\begin{flalign}\label{eq:supp-z-har-0}
\text{(HM)}&& 
\begin{split}
\lim_{c_{\text{HM}}\to 0} Z_{q,v^*,k^*}&=-\infty\\
\lim_{c_{\text{HM}}\to 0} \TIE\quad\mkern10mu &=z_{q,v,k}-C\\
&\propto z_{q,v,k},
\end{split}&&
\end{flalign}
where we use a extremely negative number $C$ to replace $-\infty$ for valid estimation of TIE. In this case, NDE is estimated as the same constant for all the answers, and TIE is dominated by $z_{q,v,k}$, which means that the language bias is not reduced. For $c_{\text{HM}}\!\to\!1$, we have
\begin{flalign}\label{eq:supp-z-har-1}
\text{(HM)}&& 
\begin{split}
\lim_{c_{\text{HM}}\to 1} Z_{q,v^*,k^*}&=\log\frac{\sigma(z_q)}{1+\sigma(z_q)}\\
\lim_{c_{\text{HM}}\to 1} \TIE\quad\mkern10mu&=\log\frac{\sigma(z_v)\cdot\sigma(z_k)\cdot(1+\sigma(z_q))}{1+\sigma(z_q)\cdot\sigma(z_v)\cdot\sigma(z_k)}.
\end{split}&&
\end{flalign}

For SUM, we have 
\begin{flalign}\label{eq:supp-z-sum}
\text{(SUM)}&&Z_{q,v^*,k^*}=\log\sigma(z_q+2c),&&
\end{flalign}
where $c\in(-\infty,+\infty)$. We approximate the limits of $Z_{q,v^*,k^*}$ and $\TIE=Z_{q,v,k}-Z_{q,v^*,k^*}$ as:
\begin{flalign}\label{eq:supp-z-sum-0}
\text{(SUM)}&& 
\begin{split}
\lim_{c\to -\infty} Z_{q,v^*,k^*}&=-\infty\\
\lim_{c\to -\infty} \TIE\quad\mkern10mu &=z_{q,v,k}-C\\
&\propto z_{q,v,k}.
\end{split}&&
\end{flalign}
Similar to HM, TIE is dominated by $z_{q,v,k}$. For $c\!\to\!+\infty$, we have:
\begin{flalign}\label{eq:supp-z-har-1}
\text{(SUM)}&& 
\begin{split}
\lim_{c\to +\infty} Z_{q,v^*,k^*}&=0\\
\lim_{c\to +\infty} \TIE\quad\mkern10mu &=z_{q,v,k}.
\end{split}&&
\end{flalign}
Also, TIE is dominated by $z_{q,v,k}$. In both cases, the language bias cannot be excluded. This analysis shows that a extremely large or small $c$ will fail to estimate NDE and TIE, and it is necessary to control the sharpness of NDE by selecting a optimal $c$. In the main paper, we use a KL-divergence in Eq. (17) to force the sharpness of NDE similar to that of TE.
\section{Implementation Details}\label{sec:supp-3}
We use the same implementation of RUBi~\cite{cadene2019rubi} for fair comparison, including feature representation, baseline architectures, and optimization.

\noindent \textbf{Image Representation}. Following the popular bottom-up attention mechanism~\cite{anderson2018bottom}, we use a Faster R-CNN based framework to extract visual features. We select top-$K$ region proposals for each image, where $K$ is fixed as 36.

\noindent \textbf{Question Representation}. Following~\cite{cadene2019murel,cadene2019rubi}, we first lowercase all the questions and remove the punctuation, and then use the pretrained Skip-thought encoder~\cite{kiros2015skip} with fine-tuning. The size of final embedding is set as 4800.

\noindent \textbf{Vision-Language Branch}. The vision-language branch consists of the image representation, question representation, and a visual knowledge encoder. The baseline models for encoding visual knowledge includes SAN~\cite{yang2016stacked}, UpDn~\cite{anderson2018bottom}, and a simplified version of the recent architecture MUREL~\cite{cadene2019murel} (S-MUREL) proposed in~\cite{cadene2019rubi}. In short, S-MUREL consists of a BLOCK~\cite{ben2019block} bilinear fusion between image and question representations for each region, and a MLP classifier composed of three fully connected layers with ReLU activations. The dimension are 2,048, 2,048, and 3,000. More details can be found in~\cite{cadene2019rubi}.

\noindent \textbf{Language-Only Branch}. The language-only branch consists of the question representation and a question-only classifier. The question-only classifier is implemented by a MLP with three fully connect layers with ReLU activations. Note that this MLP has the same structure with the classifier for vision-language branch with different parameters.

\noindent \textbf{Vision-Only Branch}. The vision-only branch is composed of the question representation and a vision-only classifier. The vision-only classifier has the same structure as the language-only classifier with different parameters.

\noindent \textbf{Optimization}. All the experiments are conducted with the Adam optimizer for 22 epochs. The learning rate linearly increases from $1.5\times 10^{-4}$ to $6\times 10^{-4}$ for the first 7 epochs, and decays after 14 epochs by multiplying 0.25 every two epochs. The batch size is set as 256.

\noindent \textbf{Datasets}. The experiments are conducted on VQA-CP~\cite{agrawal2018don} and VQA~\cite{goyal2017making} datasets. VQA-CP v1 and v2 are created by re-organizing the train and val splits of the VQA v1 and v2 datasets, respectively~\cite{agrawal2018don}. 

\section{Supplementary Experimental Results}\label{sec:supp-4}

\begin{table*}[t]
\centering
\caption{\textbf{Ablation of CF-VQA} on VQA-CP v1 test set. ``SAN/UpDn/S-MRL'' denotes the baseline VQA model. ``HM/SUM'' represents the strategies that train the ensemble model and test with only the vision-language branch following ensemble-based method~\cite{cadene2019rubi,clark2019don}. $^*$ represents the reproduced results.}
\label{tab:supp-2-1-ablation-cp1}
\centering

\scalebox{0.81}{
    \begin{tabular}{l cccc}
        \hline
        \toprule
          & All & {Y/N} &  {Num.} & {Other}\\
        \hline
        SAN$^*$ & 32.50 & 36.86 & 12.47 & 36.22 \\
        \hline
        Harmonic & 49.29 & 72.73 & \bf 20.57 & 37.51 \\
        +~CF-VQA & \bf 52.06 & \bf 80.38 & 16.88 & \bf 38.04 \\
        \hline
        SUM & 38.34 & 49.88 & \bf 15.82 & 35.91 \\
        +~CF-VQA & \bf 52.87 & \bf 84.94 & 14.85 & \bf 36.26 \\
        \bottomrule
        \hline
    \end{tabular}
}
\hfill
\scalebox{0.81}{
    \begin{tabular}{l cccc}
        \hline
        \toprule
          & All & {Y/N} &  {Num.} & {Other}\\
        \hline
        UpDn$^*$ & 37.08 & 42.46 & 12.76 & 41.50 \\
        \hline
        Harmonic & \bf 55.75 & 80.65 & \bf 24.72 & 43.46 \\
        +~CF-VQA & 55.16 & \bf 82.27 & 16.14 & \bf 43.87  \\
        \hline
        SUM & 52.78 & 78.71 & 14.30 & 42.45  \\
        +~CF-VQA & \bf 57.39 & \bf 88.46 & \bf 14.80 & \bf 43.61 \\
        \bottomrule
        \hline
    \end{tabular}
}
\hfill
\scalebox{0.81}{
    \begin{tabular}{l cccc}
        \hline
        \toprule
          & All & {Y/N} &  {Num.} & {Other} \\
        \hline
        S-MRL$^*$ & 36.68 & 42.72 & 12.59 & 40.35 \\
        \hline
        Harmonic & 53.55 & 79.38 & 17.39 & 42.38  \\
        +~CF-VQA & \bf 55.26 & \bf 82.13 & \bf 18.03 & \bf 43.49  \\
        \hline
        SUM & 49.44 & 76.49 & 16.23 & 35.90 \\
        +~CF-VQA & \bf 57.03 & \bf 89.02 & \bf 17.08 & \bf 41.27  \\
        \bottomrule
        \hline
    \end{tabular}
}
\end{table*}
\begin{table*}[t]
\centering
\caption{\textbf{Ablation of CF-VQA with the simplified causal graph} on VQA-CP v1 test set. ``SAN/UpDn/S-MRL'' denotes the baseline VQA model. ``HM/SUM'' represents the strategies that train the ensemble model and test with only the vision-language branch following ensemble-based method~\cite{cadene2019rubi,clark2019don}. $^*$ represents the reproduced results.}
\label{tab:supp-2-2-ablation-cp1-simple}
\centering

\scalebox{0.81}{
    \begin{tabular}{l cccc}
        \hline
        \toprule
          & All & {Y/N} &  {Num.} & {Other}\\
        \hline
        SAN$^*$ & 32.50 & 36.86 & 12.47 & 36.22 \\
        \hline
        Harmonic & 46.83 & 66.64 & 19.45 & 38.13 \\
        +~CF-VQA & \bf 54.48 & \bf 83.73 & \bf 22.73 & \bf 38.15 \\
        \hline
        SUM & 40.08 & 54.15 & 15.53 & \bf 35.95 \\
        +~CF-VQA & \bf 52.73 & \bf 84.64 & \bf 16.02 & 35.75 \\
        \bottomrule
        \hline
    \end{tabular}
}
\hfill
\scalebox{0.81}{
    \begin{tabular}{l cccc}
        \hline
        \toprule
          & All & {Y/N} &  {Num.} & {Other}\\
        \hline
        UpDn$^*$ & 37.08 & 42.46 & 12.76 & 41.50 \\
        \hline
        Harmonic & 54.13 & 80.60 & 15.75 & 43.24\\
        +~CF-VQA & \bf 56.19 & \bf 85.08 & \bf 16.00 & \bf 43.61\\
        \hline
        SUM & 51.20 & 74.70 & 13.61 & 42.94 \\
        +~CF-VQA & \bf 56.80 & \bf 87.76 & \bf 13.89 & \bf 43.25 \\
        \bottomrule
        \hline
    \end{tabular}
}
\hfill
\scalebox{0.81}{
    \begin{tabular}{l cccc}
        \hline
        \toprule
          & All & {Y/N} &  {Num.} & {Other} \\
        \hline
        S-MRL$^*$ & 36.68 & 42.72 & 12.59 & 40.35 \\
        \hline
        Harmonic & 54.51 & 80.82 & 17.30 & 43.29  \\
        +~CF-VQA & \bf 56.82 & \bf 86.01 & \bf 17.38 & \bf 43.63\\
        \hline
        SUM & 52.54 & 78.42 & 16.77 & \bf 41.18 \\
        +~CF-VQA & \bf 57.07 & \bf 89.28 & \bf 17.39 & 41.00 \\
        \bottomrule
        \hline
    \end{tabular}
}
\vspace{-4mm}
\end{table*}

\begin{figure*}
\centering
\includegraphics[width=\textwidth]{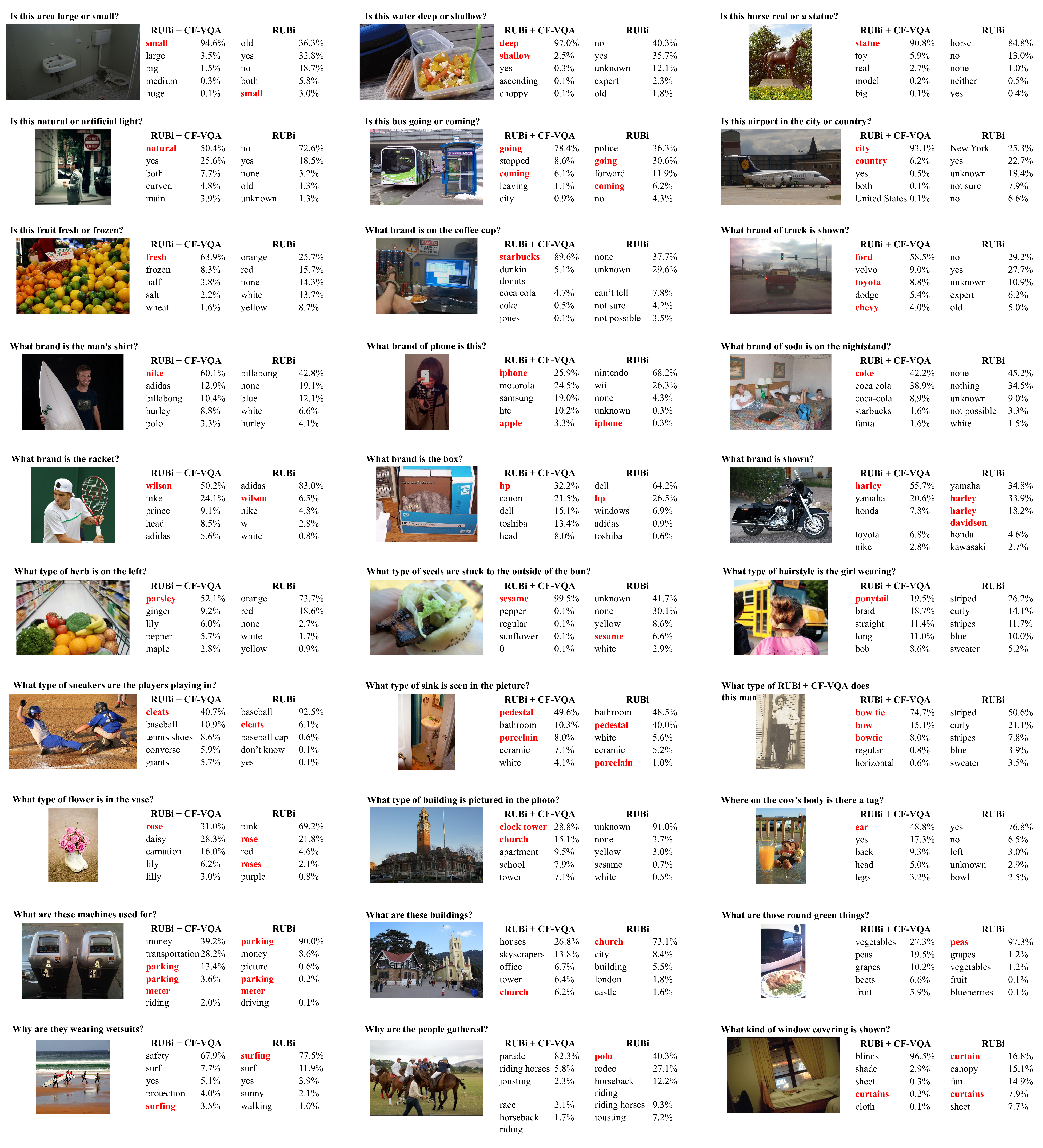}
\caption{Qualitative comparison of RUBi and RUBi+CF-VQA on VQA-CP v2 test split. Red bold answer denotes the ground-truth one.}
\label{fig:supp-rubi}
\end{figure*}

We have conducted the ablation study and compared CF-VQA with state-of-the-art methods in the main paper. In this section, we show supplementary experimental results.

\subsection{Quantitative Results}

As suggested by~\cite{teney2020value,grand2019adversarial,teney2020learning,teney2020unshuffling}, we further hold out 8,000 instances from the training set (\ie, VQA-CP v2 val) to measure the in-domain performance. Note that the results on VQA v2 val set also measure the in-domain performance. The results are given in Table~\ref{tab:supp-1-val}. Compared to GRLS~\cite{grand2019adversarial}, all of our variants outperform GRLS by large margins for both in-domain and out-of-distribution (OOD) settings. Compared tp GradSup~\cite{teney2020learning}, CF-VQA (HM) achieves better results on both VQA-CP val set and test set. Compared to RandImg~\cite{teney2020value}, CF-VQA (SUM) achieves competitive results on VQA-CP v2 test set, and outperforms RandImge on in-domain settings by over 3\%. These results demonstrate that CF-VQA not only effectively reduces language bias, but also performs robustly.

Table~\ref{tab:supp-2-1-ablation-cp1} shows the ablation study on VQA-CP v1 test split. As shown in Table~\ref{tab:supp-2-1-ablation-cp1}, CF-VQA is general to both \textit{baseline VQA architectures} and \textit{fusion strategies}, which is also demonstrated by the results on VQA-CP v2. Table~\ref{tab:supp-2-2-ablation-cp1-simple} shows the ablation study on VQA-CP v1 test split using the simplified causal graph. Similarly, CF-VQA achieves significant improvement for different baseline VQA architectures and fusion strategies.

\subsection{Qualitative Results}

Figure~\ref{fig:supp-rubi} illustrates examples to show how CF-VQA improves RUBi by simply replacing natural indirect effect with total indirect effect for inference following Algorithm~\ref{alg:supp-rubi}. The examples show that CF-VQA benefits from language context, \eg, ``large or small'', ``deep or shallow'', and ``real or a statue'' in the first row. Some failure cases are shown in the last two rows. First, CF-VQA may tend to generate broad answers, \eg, ``houses'' v.s ``church'', and ``vegetables'' v.s ``peas''. Second, CF-VQA may ignore visual content like traditional likelihood strategy. Therefore, there remains the challenge about how to balance visual understanding and language context.

{\small
\bibliographystyle{ieee_fullname}
\bibliography{main}
}

\end{document}